\pdfoutput=1

\documentclass[11pt]{article}

\usepackage{ACL2023}

\usepackage{times}
\usepackage{xspace}
\usepackage{latexsym}
\usepackage{float}
\usepackage{booktabs}
\usepackage[T1]{fontenc}

\usepackage[utf8]{inputenc}

\usepackage{microtype}

\usepackage{inconsolata}
\usepackage{subcaption}

\usepackage{adjustbox}
\usepackage{xcolor}
\usepackage{float}
\usepackage{siunitx}
\usepackage{bm}
\usepackage{amsmath}
\usepackage[capitalize,noabbrev]{cleveref}
\usepackage{listings}
\usepackage{array}
\usepackage{multirow}
\usepackage{tabularx}
\usepackage[frozencache=true,cachedir=minted-output]{minted}
\usepackage{color, colortbl,booktabs}
\usepackage{diagbox}

\newcolumntype{H}{>{\setbox0=\hbox\bgroup}c<{\egroup}@{}}
\newcolumntype{Z}{>{\setbox0=\hbox\bgroup}c<{\egroup}@{\hspace*{-\tabcolsep}}}

\title{Sparse Logit Sampling: Accelerating Knowledge Distillation in LLMs}

\author{Anshumann* \and Mohd Abbas Zaidi* \and Akhil Kedia* \and Jinwoo Ahn \\
 {\bf Taehwak Kwon \and  Kangwook Lee \and Haejun Lee \and Joohyung Lee} \\
  Samsung Research, Seoul \\
  \texttt{\{anshu.mann, abbas.zaidi, akhil.kedia, jinwoo.ahn, taehwak.kwon\}@samsung.com}}

\begin{document}
\maketitle
\begin{abstract}
Knowledge distillation can be a cost-effective technique to distill knowledge in Large Language Models, if the teacher output logits can be pre-computed and cached. However, successfully applying this to pre-training remains largely unexplored. In this work, we prove that naive approaches for sparse knowledge distillation such as caching Top-K probabilities, while intuitive, provide biased estimates of teacher probability distribution to the student, resulting in suboptimal performance and calibration. We propose an importance-sampling-based method `Random Sampling Knowledge Distillation', which provides unbiased estimates, preserves the gradient in expectation, and requires storing significantly sparser logits. Our method enables faster training of student models with marginal overhead ($<10\%)$ compared to cross-entropy based training, while maintaining competitive performance compared to full distillation, across a range of model sizes from $300$M to $3$B.
\end{abstract}

\section{Introduction}

Distilling the knowledge from a larger teacher into a smaller student ~\citep{DistillingKnowledgeNeural} has been successfully used to train more efficient and stronger models across a range of applications \citep{fukuda2017efficient, jiao2019tinybert, ahn2019variational, tian2019contrastive, sanh2020distilbertdistilledversionbert, bergmann2020uninformed, zhao2022decoupled, SurveyKnowledgeDistillation}. As Large Language Models (LLMs) reach increasing adoption, Knowledge Distillation has also been applied to improve smaller LLMs \citep{LLMPruningDistillation, CompactLanguageModels, MiniPLMKnowledgeDistillation, MiniLMv2MultiHeadSelfAttention, MiniLLMKnowledgeDistillation, PerformanceGuidedLLMKnowledge, CrossTokenizerDistillationUniversal, LionAdversarialDistillation}. \def\thefootnote{*}\footnotetext{Equal Contribution. Source code is available \href{https://github.com/akhilkedia/RandomSamplingKD}{here}.}\def\thefootnote{\arabic{footnote}}

\begin{figure}[t]
     \centering
     \includegraphics[width=\linewidth]{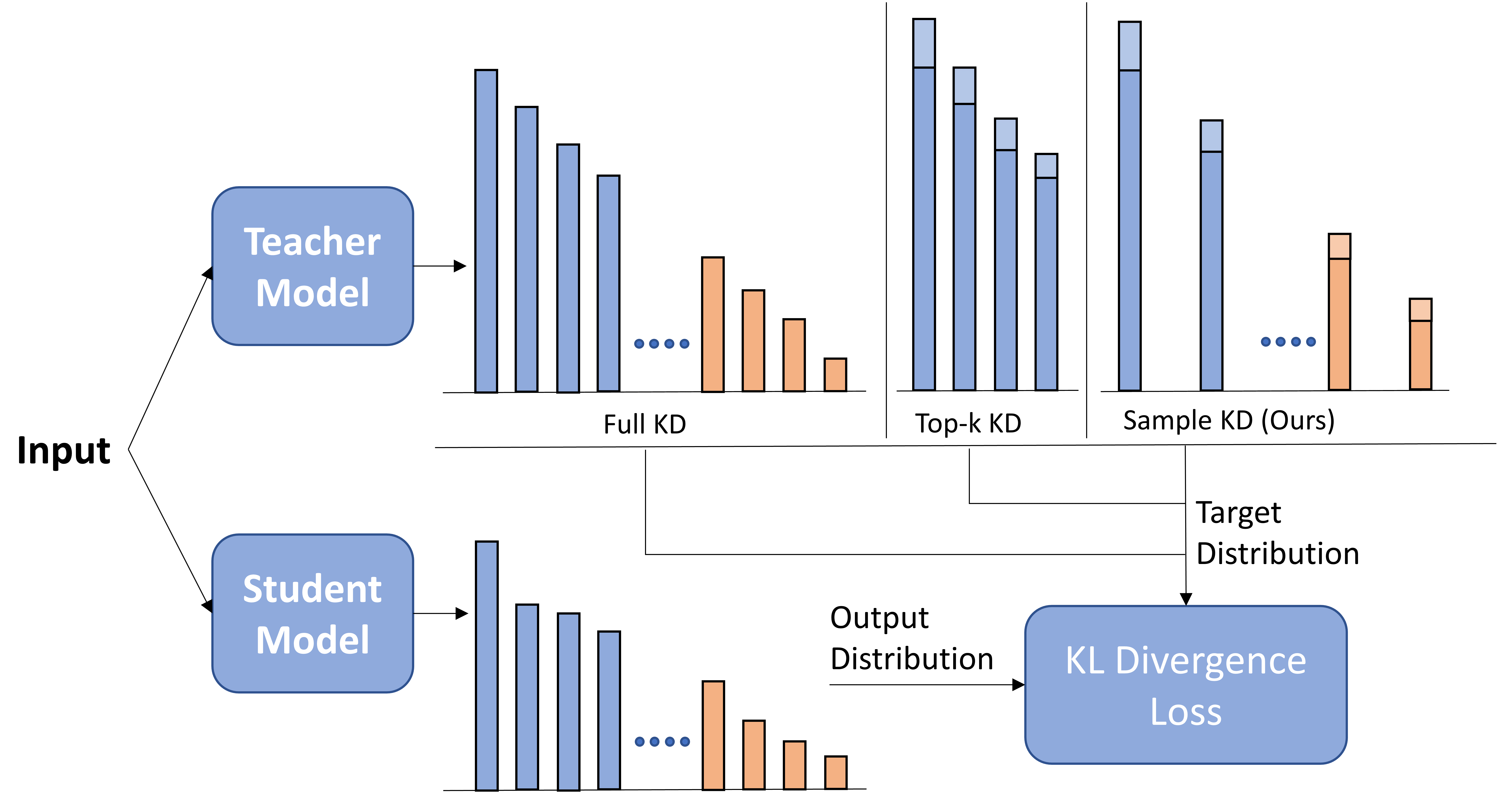}
     \caption{Sparse Knowledge Distillation Pipeline}
     \label{fig:main}
 \end{figure}

Two common categories of Knowledge Distillation are distribution matching, where the teacher's final logits or output distribution are learned, and representation matching, where intermediate-layer representations are distilled~\citep{FDivergenceMinimizationSequenceLevel}. In this work, we focus on the former, in a \textit{offline logits} setting, where the logits from the teacher are pre-computed and cached, prior to training the student.  

Particularly for LLMs, this setting has several advantages -- The larger, more expensive teacher only needs to run once, and the saved representations can then be used to train a family of smaller models of various sizes. Teacher inference can be done on cheaper compute resources without fast multi-node interlinks, and the student can be trained on smaller clusters. Cluster size is further reduced by eliminating the memory footprint of the teacher. Lastly, this makes smaller-scale design experiments and ablations feasible by eliminating the constant large overhead of running the teacher model repeatedly for each experiment or training. 

While this is often done for post-training~\citep{FIRSTTeachReliable} or for dataset generation/filtering~\citep{MiniPLMKnowledgeDistillation, FDivergenceMinimizationSequenceLevel, TextbooksAreAll}, extending this to pre-training is challenging. In contrast to vanilla pre-training, knowledge distillation requires the information-dense soft targets (teacher probabilities) to be stored. Due to the large vocabulary size of modern LLMs, naively storing all of these probabilities is infeasible (e.g., requiring $128$ PetaBytes of storage for $1$T tokens for Llama3~\citep{grattafiori2024llama3herdmodels}). Instead, sparse knowledge distillation approaches store an efficient Top-K subset of logits from the teacher's distribution~\citep{DistillationTokensAre, PretrainingDistillationLarge}. However, these methods still require a large number of logits ($6400$) to be stored, or even observe a \textit{fall} in model performance~\citep{PretrainingDistillationLarge}.


In this work, via theoretical proofs, cross-validated by empirical analysis, we show that the performance drop in Top-K methods stem from two primary causes - 1) Top-K provides a biased estimator of the teacher's probability distribution, and 2) It fails to expose the tail of teacher's distribution to the student model. These result in the student learning a scaled-up and mis-calibrated distribution of the teacher probability.

\begin{figure*}[t]
\begin{minipage}[ht]{0.32\linewidth}
\centering
    \begin{subfigure}[b]{\textwidth}
         \centering
         \includegraphics[width=\linewidth]{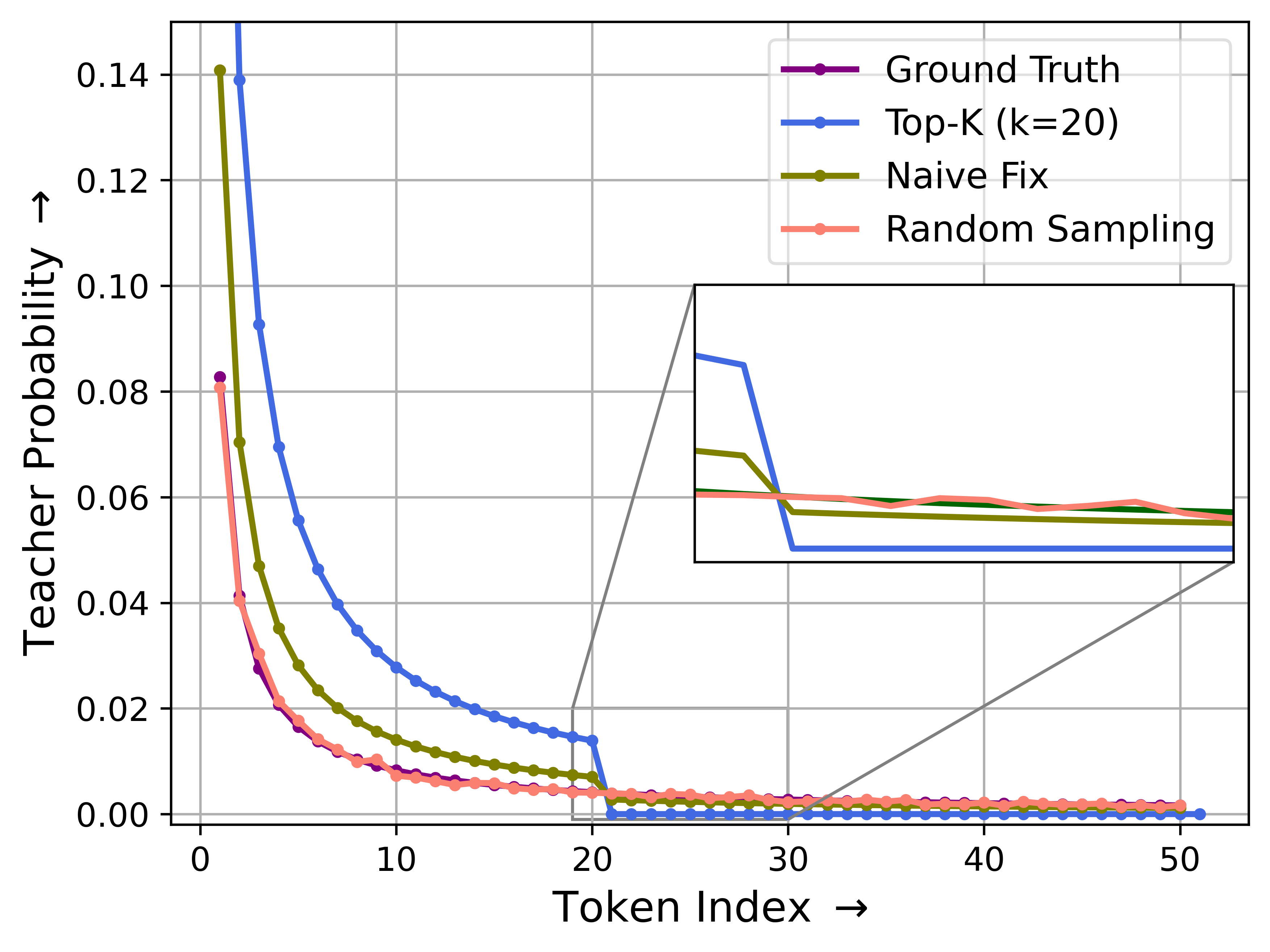}
         \caption{Visualizing Target Probabilities}
         \label{fig:toy-sampling-compare}
    \end{subfigure}
\end{minipage}
\hfill
\begin{minipage}[ht]{0.32\linewidth}
\centering
    \begin{subfigure}[b]{\textwidth}
        \centering
     \includegraphics[width=\linewidth]{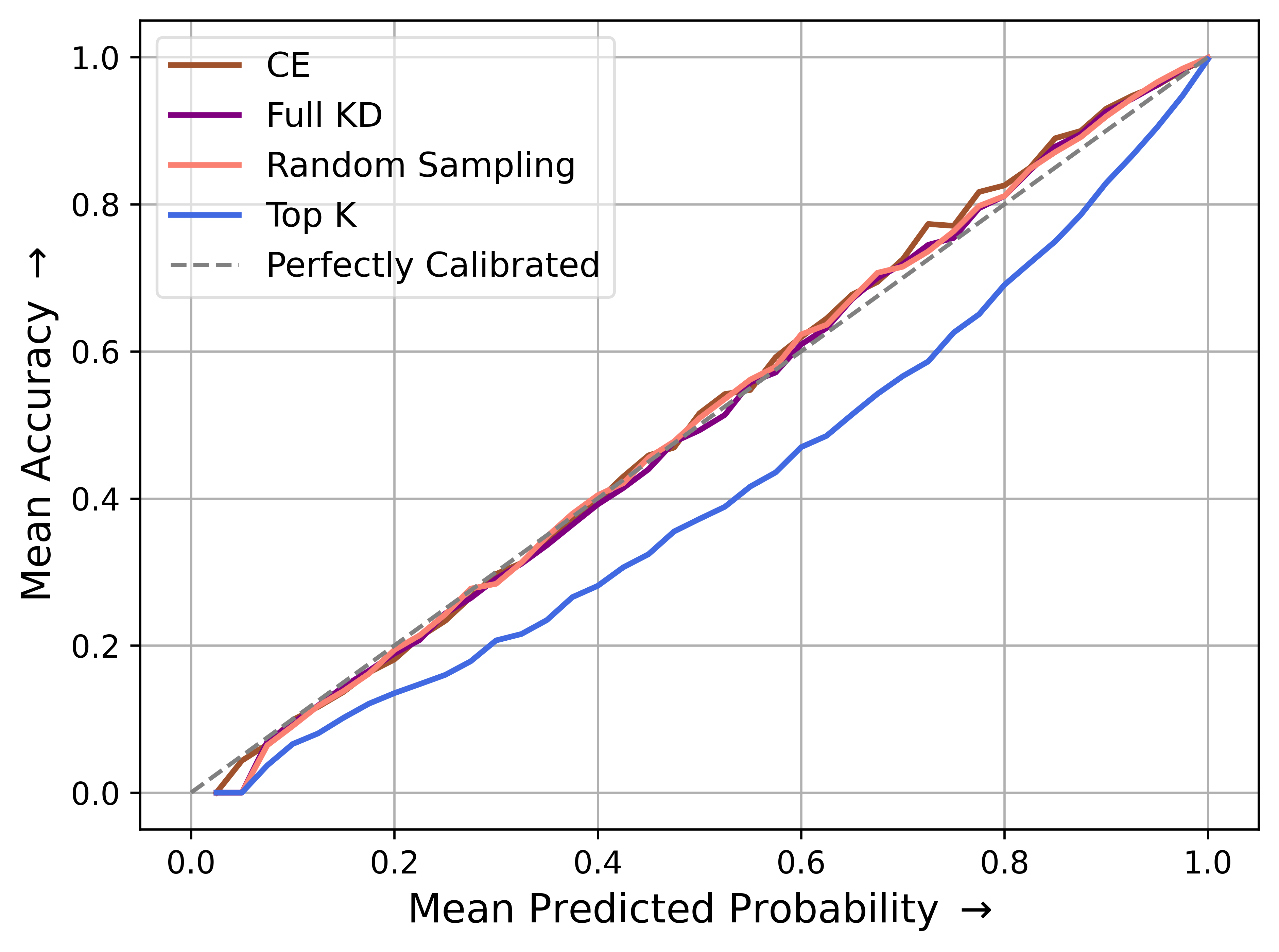}
     \caption{Calibration on Synthetic Classes}
     \label{fig:toy-calibration-compare}
    \end{subfigure}
\end{minipage}
\hfill
\begin{minipage}[ht]{0.32\linewidth}
\centering
    \begin{subfigure}[b]{\textwidth}
        \centering
     \includegraphics[width=\linewidth]{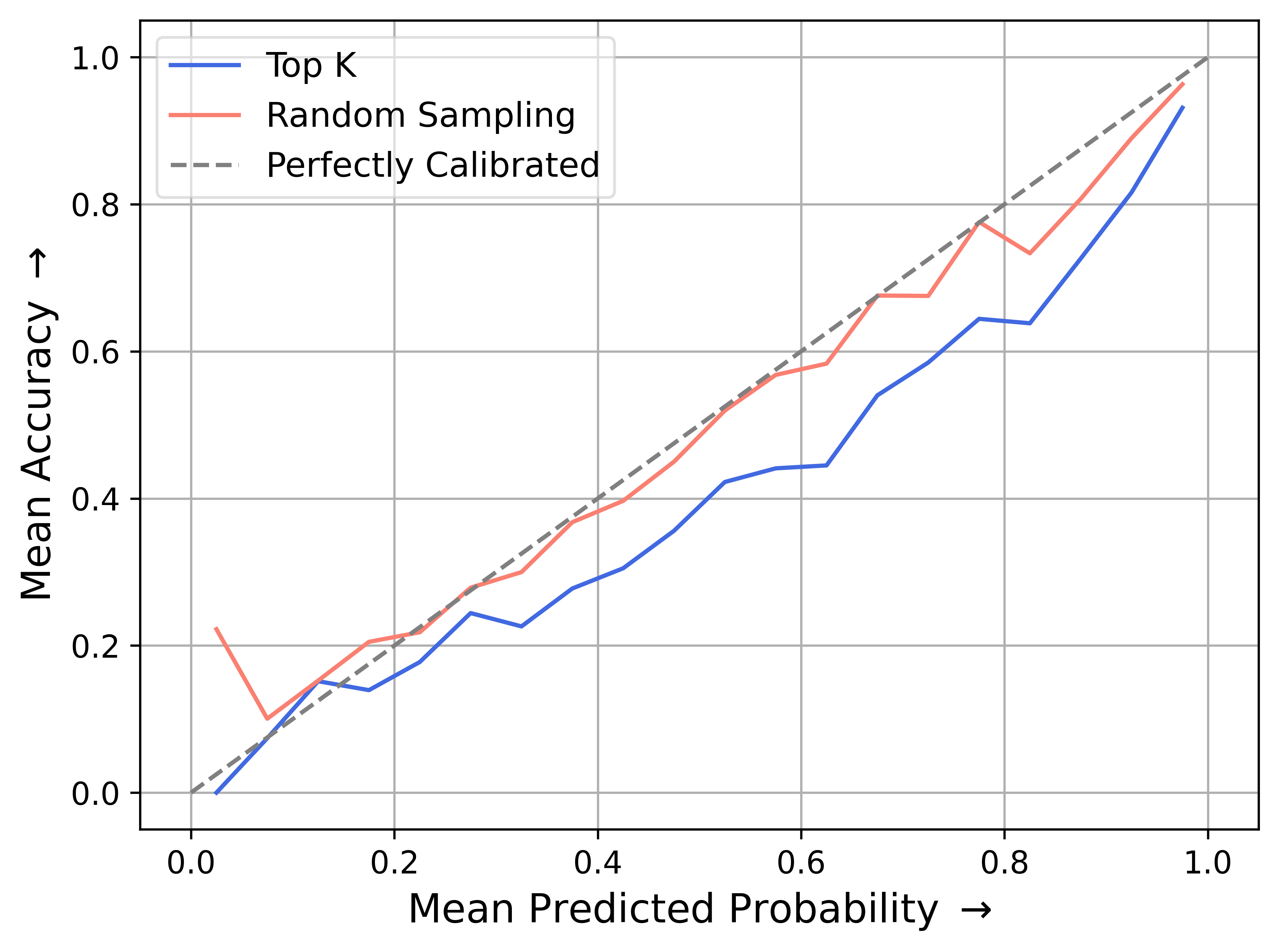}
     \caption{Calibration on CIFAR-100}
     \label{fig:cifar-calibration}
    \end{subfigure}
\end{minipage}
\caption{Comparing different sparse KD methods on synthetic examples (refer to \cref{sec:synthetic}).}
\label{fig:toy}
\end{figure*}

We rectify both of these issues by instead utilizing importance sampling~\citep{AdvancesImportanceSampling} to randomly sample from the teacher's distribution. We show that our proposed Knowledge Distillation approach -- 1) Provides an unbiased estimate of the teacher's probability distribution, 2) Preserves the gradient in expectation compared to full distillation, and 3) Eliminates the overhead of running the teacher inference, while maintaining model performance to full distillation, using extremely limited storage.

\section{Top-K Knowledge Distillation}

For storing KD logits, previous studies~\citep{DistillationTokensAre, PretrainingDistillationLarge, FIRSTTeachReliable} have proposed to replace the full teacher distribution $\mathbf{t}$ in knowledge distillation with a sub-sampled version $\mathbf{t^s}$. The most intuitive way is to use only the top $K$ probability values from the teacher ("Top-K KD"), specifically $ t^s_i=t_i, i \in K$, and $t^s_i=0$ otherwise, where $t_i$ are the probabilities of the token $i$ in $\mathbf{t}$. Note that $\sum{t^s_i} \neq 1$.

Theoretically, selecting the top $K$ tokens results in the least error from the teacher distribution for a single token~(\cref{topk_error_proof}). This may be combined with "Top-p" which  dynamically adjusts $K$ to only keep a fixed probability mass $p$.

\subsection{How Does Top-K perform compared to FullKD?}

To study Top-K KD, we pre-train multiple LLaMA style $300$M student models, while varying the number of probabilities used $K$. We train on web data using a well pre-trained $3$B teacher (full hyper-parameters in \cref{hparams_300M}), using forward KL-Divergence loss. As a baseline, we use a model trained with only Cross Entropy loss ("CE"), and as a ceiling, a model trained using the entire teacher distribution ("FullKD") to compare student performance on language modeling tasks.

As seen in the table \cref{table: Vanilla Top-K}, Top-K training lags behind the FullKD performance on the language modeling task. If a small number of Top-K tokens ($<25$) are used, the student loss is worse than just than using CE loss without any distillation -- Only after $300$ tokens does the model performance start reaching close to FullKD. Using Top-p allows for the use of fewer tokens, but performance is still only $47\%$ of FullKD. 

We also measure the Expected Calibration Error~\citep{CalibrationModernNeural} ("ECE") of these models, as prior works~\citep{FIRSTTeachReliable} have shown that calibration is strongly correlated with performance. Even though our teacher model is almost perfectly calibrated, we find that models trained with Top-K are strongly mis-calibrated, with calibration worsening as number of tokens (K) is being reduced. Models trained using CE and FullKD are almost perfectly calibrated, as has also been previously observed~\citep{CalibrationLargeLanguage, FIRSTTeachReliable, UnderstandingCalibrationTransfer}.

\begin{table}[H]
\begin{center}
\begin{adjustbox}{max width=0.7\linewidth}
\begin{tabular}{rS[table-format=1.2, round-mode=places, round-precision=2]S[table-format=-3\%, round-mode=places, round-precision=0]S[table-format=2.1, round-mode=places, round-precision=1]}
\toprule
    \textbf{Unique} & \textbf{LM}  & \textbf{\% CE to} & \textbf{ECE}  \\
    \textbf{Tokens} & \textbf{Loss $\downarrow$}  & \textbf{FullKD $\uparrow$} & \textbf{\%$ \downarrow$}  \\
    \midrule
    \text{CE}  & 2.80668 & 0\% & 1.154 \\
    \midrule
    $3$  & 3.04272 & -395.3768844\% & 10.559 \\
    $5$  & 2.95769 & -252.9480737\% & 7.677 \\
    $12$  & 2.86562 & -98.72696817\% & 4.723 \\
    $25$  & 2.81908 & -20.77051926\% & 3.18 \\
    $50$  & 2.80378 & 4.857621441\% & 2.204 \\
    *$50$   & 2.77848 & 47.2361809\% & 1.681 \\
    $57$  & 2.78745 & 32.21105528\% & 2.002 \\
    $100$  & 2.77411 & 54.5561139\% & 1.129 \\
    $300$  & 2.76044 & 77.45393635\% & 1.515 \\
    \midrule
    \text{FullKD}  & \bfseries 2.74698 & \bfseries 100\% & \bfseries 0.73 \\
    \bottomrule
\end{tabular}
\end{adjustbox}
\end{center}
\caption{Vanilla Top-K KD. The row *$50$ uses Top-p $0.98$ with $K=100$. `\% CE to FullKD' refers to the \% gap covered between CE and FullKD models.}
\label{table: Vanilla Top-K}
\end{table}

\subsection{Top-K KD Analysis}

In this section, we demonstrate fundamental problems with Top-K methods.

\subsubsection{Up-scaled Teacher Probabilities}

\paragraph{Synthetic Toy Distribution:} When only the Top-K values are kept from the teacher distribution, the probabilities of the top tokens are inevitably scaled-up compared to the original, as the probabilities must be normalized to sum to $1$. We illustrate this in \cref{fig:toy-sampling-compare} (see \cref{toy_prob_pseduo} for pseudo-code), where we simulate a synthetic distribution following a Zipf distribution~\citep{zipf}. Similar bias was also observed in previous research~\citep{BiasCrossEntropyBasedTraining}.

\paragraph{Gradients from KL-Divergence:} When using KL-Divergence loss with Top-K KD, the non Top-K tokens are pushed to probability $0$ due to restriction of the target distribution to Top-K probabilities. This happens even if one does not normalize the Top-K teacher probabilities. The backward gradients result in the student effectively learning an up-scaled version of the teacher probabilities as targets, with the remaining probability divided among the Top-K tokens. Specifically, if $p_i$ and $t_i$ are the student and teacher probabilities for the $i^{th}$ token, the gradients for the $i_{th}$ logit $x_i$ in FullKD are:
\begin{align}
    \frac{\partial L}{\partial x_i} &= p_{_i} - t_i
\end{align}
But for Top-K KD, as we prove in \cref{naive_upscales}, the gradients are:
\begin{align}
    \frac{\partial L}{\partial x_i} &= (\sum_{j\in K} t_j). {p_{_i}} - t_i
\end{align}

\begin{figure*}[t]
\begin{minipage}[ht]{0.32\linewidth}
\centering
    \begin{subfigure}[b]{\textwidth}
        \centering
        \includegraphics[width=\linewidth]{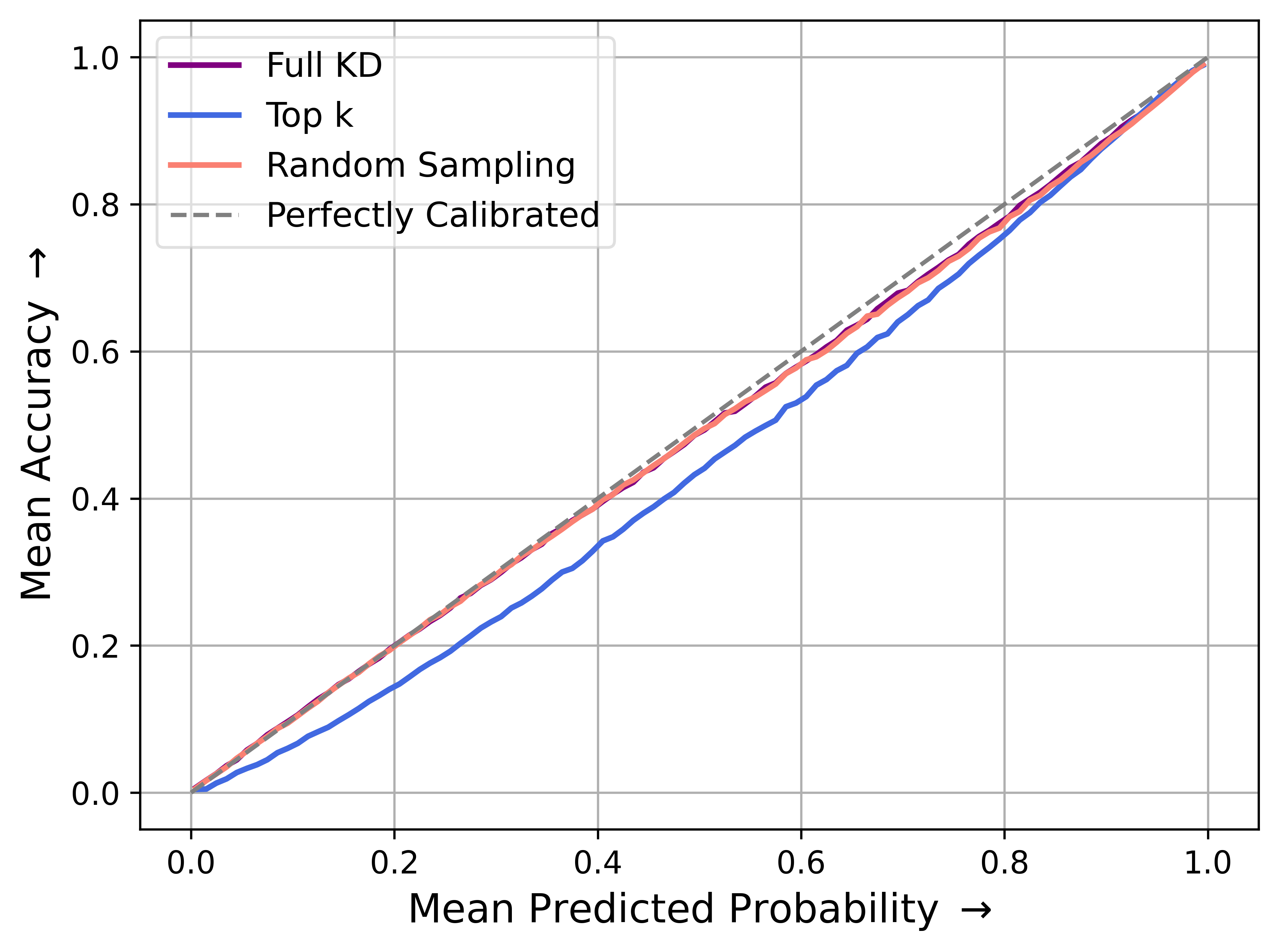}
        \caption{Calibration of Pre-trained models}
        \label{fig:calib_gauss}
    \end{subfigure}
\end{minipage}
\hfill
\begin{minipage}[ht]{0.32\linewidth}
\centering
    \begin{subfigure}[b]{\textwidth}
        \centering
        \includegraphics[width=\linewidth]{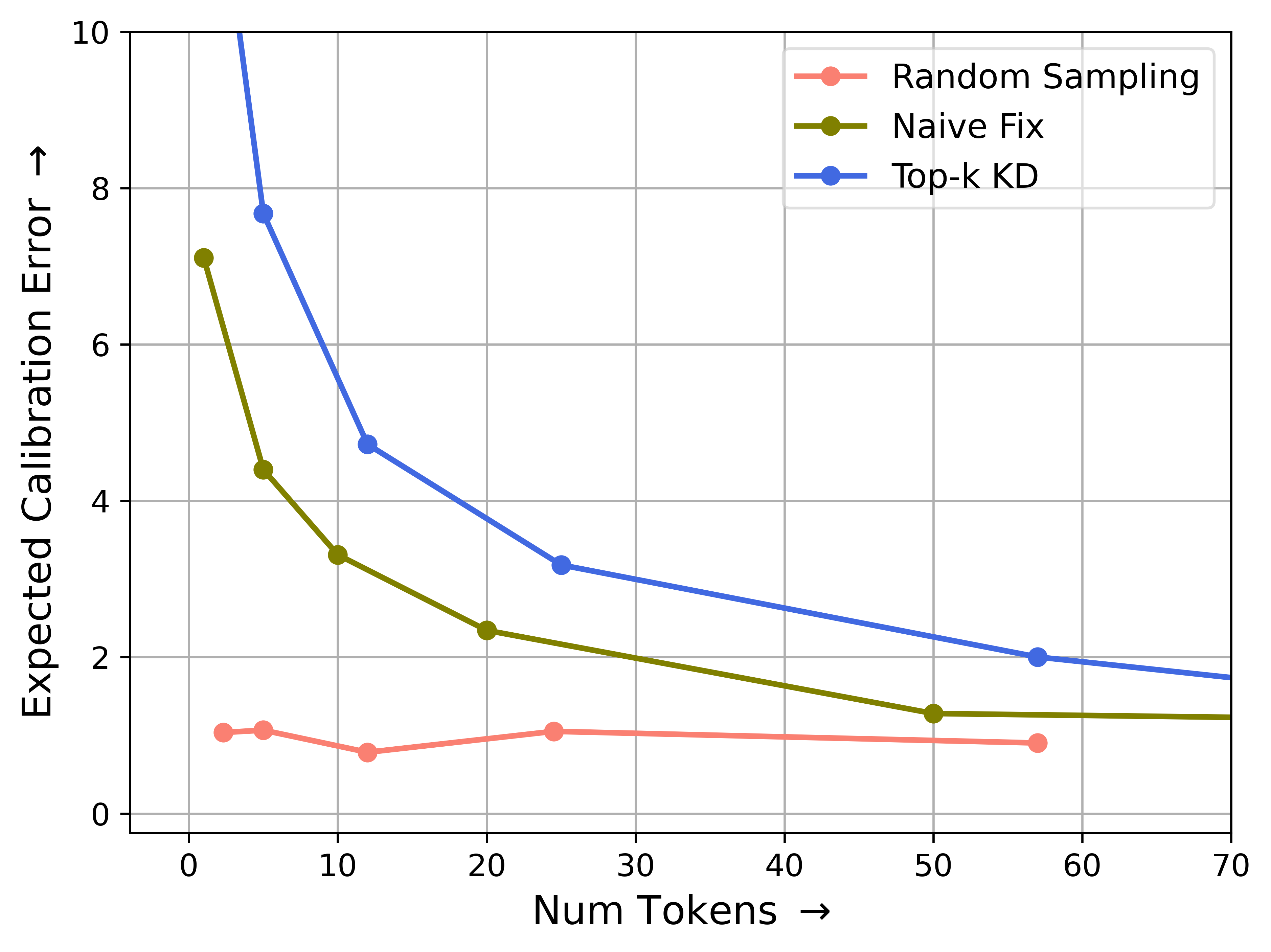}
        \caption{Expected Calibration Error}
        \label{fig:ece-compare}
    \end{subfigure}
\end{minipage}
    \hfill
\begin{minipage}[ht]{0.32\linewidth}
\centering
    \begin{subfigure}[b]{\textwidth}
        \centering
        \includegraphics[width=\linewidth]{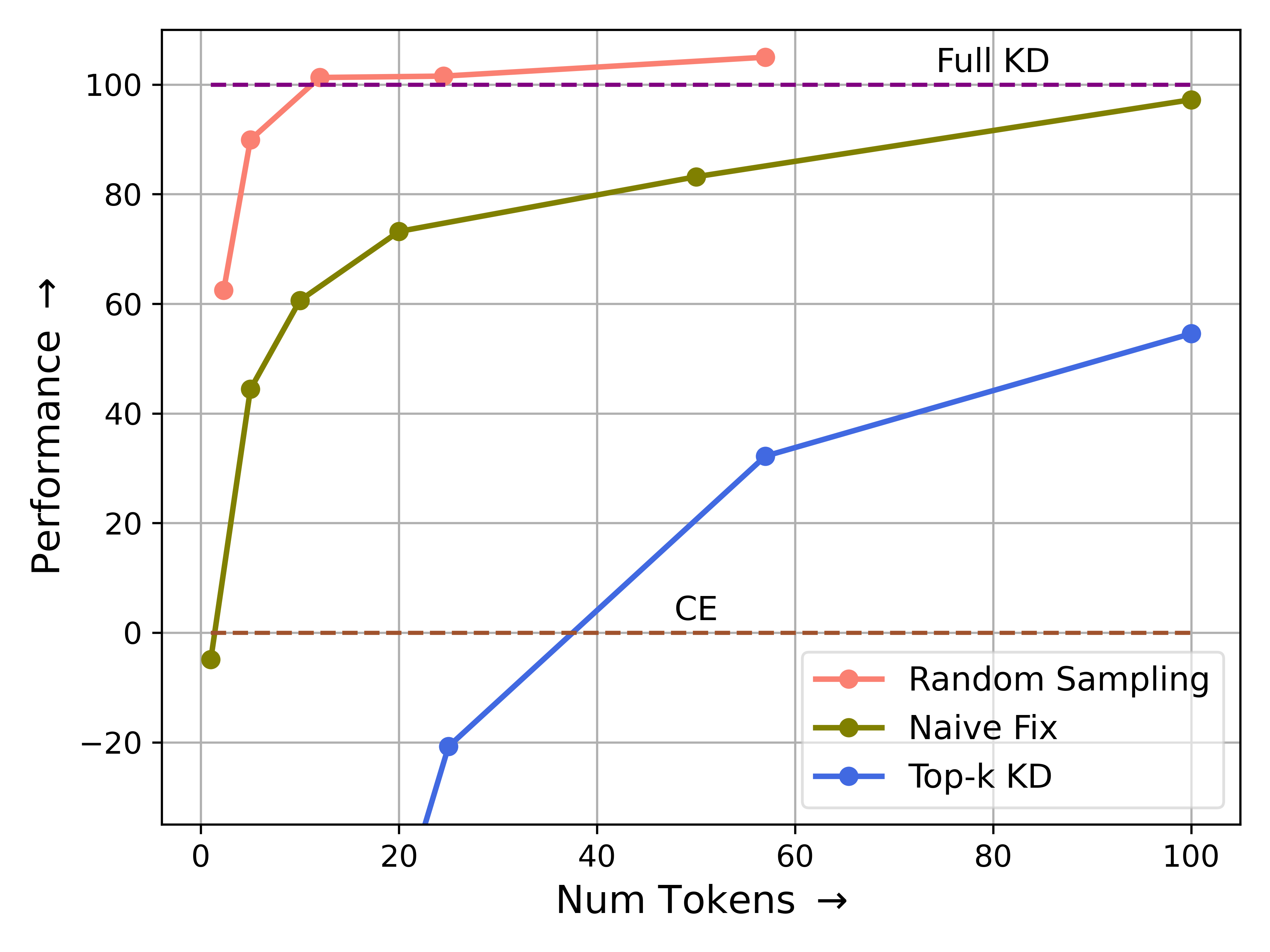}
        \caption{LM Loss Performance}
        \label{fig:kd-compare}
    \end{subfigure}
\end{minipage}
\caption{Comparing different sparse KD methods on Language Modeling Pre-Training}
\label{fig:overall-comparison}
\end{figure*}

The student will hence be over-confident in the Top-K tokens, and under-confident for the remaining tokens (\cref{naive_upscales}). This over-confidence for the top tokens is indeed what we observe with top-K pre-training for LLMs (\cref{fig:calib_gauss}), causing the calibration error in \cref{table: Vanilla Top-K}, which worsens as $K$ is decreased. Other works~\citep{distillation_scaling_laws} have also observed this top-K bias and mis-calibration, while finding the full teacher distribution to be unbiased. 

\paragraph{Synthetic Classification Task:} This calibration error can even be observed in a very simple synthetic classification task (similar to \citealp{NotBlindlyImitate}), where we train a toy $3$-layer MLP for classifying random points with Gaussian noise around class means in $128$-dimensional space (see \cref{toy_example_pseduo} for pseudo-code). As seen in \cref{fig:toy-calibration-compare}, Top-K KD leads to over-confident models, whereas CE and FullKD are almost perfectly calibrated. The same effect is observed when training a toy ResNet~\citep{resnet} model on CIFAR-100~\citep{cifar100} dataset, as shown in \cref{fig:cifar-calibration}.

Hence, we cannot apply KL-Divergence loss on the Top-K target distribution without explicitly handling the remaining probability.

\subsubsection{Missing Tail Information}

However, only handling the problem of up-scaled teacher probabilities is not sufficient to fully recover the performance~(\cref{label_smoothing,ghost_token_desc}). In contrast to FullKD training, which utilizes the full distribution, Top-K KD discards the tail information which has been shown to be crucial for model performance~\citep{ModelsCollapseTrained}. For rare ground truth tokens which fall in the tail of the teacher distribution, Top-K KD throws away the ground truth, providing a poor training signal compared to CE training. The tail, even though it contains a small probability mass, contains useful information and needs to somehow be learned.

\section{Partial Empirical Solutions}

In this section, we first discuss several empirical solutions to the problems discussed above. We apply these fixes to Top-K KD, and provide the corresponding results in \cref{table: naive fix}. 



\subsection{Label Smoothing}
\label{label_smoothing}
A straightforward solution is to distribute the residual probability over all the classes equally. Here, residual probability refers to $1-p$ where $p$ is the sum of the probabilities of the top-K tokens from the teacher's probability distribution. While this fixes the calibration error, smoothing leads to significant degradation in the performance compared to Top-K KD~(\cref{table: naive fix}). This is expected since real-world token probabilities are not uniformly distributed and are instead hyperbolic (Zipf~\citealp{zipf}). While some studies show benefits of using smoothing~\citep{StatisticalPerspectiveDistillation}, other works~\citep{KnowledgeDistillationApprox, FIRSTTeachReliable} also find that label smoothing under-performs in KD.

\begin{table}[H]
\begin{center}
\begin{adjustbox}{max width=\linewidth}
\begin{tabular}{lS[table-format=1.2, round-mode=places, round-precision=2]S[table-format=1.2, round-mode=places, round-precision=2]S[table-format=-2, round-mode=places, round-precision=0]S[table-format=.1, round-mode=places, round-precision=1]S[table-format=2.1, round-mode=places, round-precision=1]}
\toprule
    \textbf{Method} & \textbf{Top-k} & \textbf{New LM}  & \textbf{\% CE to} & \textbf{ECE} & \textbf{0-shot} \\
    \textbf{} & \textbf{Loss $\downarrow$} & \textbf{Loss $\downarrow$}  & \textbf{FullKD $\uparrow$} & \textbf{\% $\downarrow$} & \textbf{Score $\uparrow$} \\
    \midrule
    CE & 2.80668 & \text{-} & 0 & 1.154 & 40.4 \\
    \midrule
    Smoothing & 2.80378  &  2.85019  &  -72.88107203 & 0.37 & 41.2 \\
    Ghost Token & 2.80378 &  2.77134  &  59.1959799 & 0.38 & 42.9 \\
    \midrule
    \multicolumn{6}{c}{\centering \textit{Naive Fix: Remaining Probability to Ground Truth}} \\ 
    Top-k $1$  &  3.37382 &  2.80958  &  -4.857621441 & 7.11 & 41.3 \\
    Top-k $5$  & 2.95769  &  2.78016  &  44.42211055  & 4.4 & 42.4 \\
    Top-k $10$ &  2.88079 &  2.77049  &  60.61976549  & 3.31 & 42.4 \\
    Top-k $20$ & 2.83065  &  2.76298  &  73.19932998 & 2.343 & 42.9 \\
    Top-k $50$ & 2.80378  &  2.75702  &  83.18257956 & 1.279 & 42.8 \\
    Top-k $100$ & 2.77411  &  2.74863  &  97.2361809 & 1.158 & 43 \\
    \midrule
    FullKD & \bfseries 2.7369 & \text{-} & \bfseries 100 & 0.73 & 42.1  \\
    \bottomrule
\end{tabular}
\end{adjustbox}
\end{center}
\caption{Naive Fixes for Top-K KD. Smoothing (Label Smoothing) and Ghost Token use $50$ tokens.}
\label{table: naive fix}
\end{table}

\subsection{Ghost Token}
\label{ghost_token_desc}
Another method to handle residual probability would be to create a "ghost token" which takes up the accumulated probabilities of non Top-K tokens for both the teacher and the student. We compute loss on the $K$ top tokens, between predicted probabilities $p_i$ and target $t^s_i=t_i$, and on the "ghost token" with probability $p_\text{ghost} = 1-\sum_{i\in K} p_i$ and target $t^s_\text{ghost} = 1-\sum_{i\in K} t_i$.

With the ghost token, the Top-K tokens receive the same gradient as FullKD, while the remaining tokens receive gradients proportional to the student confidence~(\cref{ghost_token}). This significantly improves both the LM loss and calibration (\cref{table: naive fix}). However, the performance is still worse compared to FullKD -- indicating that explicit supervision in the tail is essential to bridge the performance gap. 

\subsection{Naive Fix}
A trivial candidate for the residual probability of the teacher is the ground truth itself. We label this method as "Naive Fix", where the probability of the target token is adjusted to ensure that the target probabilities sum up to $1$. One can expect that this will result in probabilities more aligned to the real target~(\cref{fig:toy-sampling-compare}). This method significantly improves both performance and calibration error \cref{table: naive fix}, however, it still requires $100$ tokens to achieve performance comparable to FullKD.

The gradients for the logits are linked to the target teacher probability (\cref{sec:sfotmax-kld-grad}  - \cref{eq:loss_grad_general}). The methods above are either biased estimators of the teacher probability distribution, and/or lack adequate supervision in the tail.

\section{Proposed Method: Random Sampling KD}
\label{sec:proposed}

We propose a theoretically motivated method ``Random Sampling KD'',  which overcomes all the drawbacks of the previous approaches. Given a teacher probability distribution $\mathbf{t_{full}}$ for each token $i$ in the vocab $V$, unlike Top-K which truncates the teacher distribution, our method randomly samples tokens from teacher distribution.

\paragraph{Motivation} For a given probability distribution $t(x)$, importance sampling~\citep{AdvancesImportanceSampling} allows us to obtain unbiased estimates of a function $f(x)$, by sampling from a different proposal distribution $q(x)$, and reweighing the samples using the likelihood ratio $t(x)/q(x)$.
\[
\resizebox{\columnwidth}{!}{$\displaystyle
E[f(x)] = \int f(x) t(x) dx = \int f(x) \frac{t(x)}{q(x)} q(x) dx
$}
\]
If the proposal $q(x)=0$ at any $x$ where $t(x) \neq 0$ (e.g., Top-K), then the estimate is no longer unbiased. While any non-zero proposal distribution $q(x)$ can be used to obtain an unbiased estimate, under certain constraints, the proposal distribution with the lowest variance $q^*(x)$ is of the form $q^*(x) \propto t(x) |f(x)|$~\cite{Goodfellow-et-al-2016-Importance}. Motivated by these findings, we explore $q(x) = {t(x)}^{\tau}$ as a proposal distribution, where $\tau$ is the sampling temperature.

\paragraph{Sampling Distribution} We sample tokens from $\mathbf{t_{full}}$, using the proposal distribution $\mathbf{q}=\mathbf{t^{\tau}_{full}}$, for a fixed number of rounds $N$. Each occurrence of a token $i$ is assigned a likelihood ratio $\frac{t_i}{q_i}$. Empirically, we find that for $0.8<\tau<1.2$, performance does not vary significantly~(\cref{table: Random Sampling Temperature}). We hence use $\tau=1$, simply sampling $N$ token ids from $1$ to $V$ (with replacement) with probability $\mathbf{t_{full}}$.

\paragraph{Obtaining Sampled Probabilities} For each token, the likelihood ratio of each sample is added, and then normalized to obtain the sub-sampled target probability distribution $\mathbf{t^s}$. For $\tau=1$, the likelihood ratio is simply $1$, and $t^s_i$ is then $\frac{c_i}{N}$, where $c_i$ is the count of occurrences of each token $i$ in $N$ samples. This will be very sparse, with maximum $N$ non-zero probabilities, and significantly less than $N$ in practice~(\cref{sec:sampled_vs_unique}).

\paragraph{Loss Calculation} We use forward KL divergence between non-zero $\mathbf{t^s}$ and student predictions $\mathbf{p}$, $\sum{t^s_i}log{\frac{t^s_i}{p_i}}$. For $\tau=1$, this may also be viewed as the sum of cross entropy loss between each sampled token and the student predictions. 

This sub-sampled teacher distribution $\mathbf{t^s}$ can be stored/cached on disk and re-used across multiple experiments. The above gives us our final method, `Random Sampling KD'.

\section{Analysis of Random Sampling KD}

\subsection{Calibration}
\label{res_calibration}

The toy distribution~(\cref{fig:toy-sampling-compare}) demonstrates that our method correctly estimates teacher distribution by providing an unbiased probability estimates,  It achieves perfect calibration mirroring FullKD in the synthetic classification tasks~(\cref{fig:toy-calibration-compare}), in toy classification on CIFAR-100~(\cref{fig:cifar-calibration}) and in LLM pre-training~(\cref{fig:calib_gauss}). 

As compared to the other KD methods discussed above, models trained with Random Sampling KD are much better calibrated, and using fewer tokens does not hurt the calibration~(\cref{fig:ece-compare}).

\subsection{Gradient Similarity}
\label{sec:gradient}

In \cref{sec:grads_preserved}, we prove that random sampling preserves the expected gradients at the logits when compared to FullKD. To further verify this empirically, we measure the gradients of the parameters of a $300$M model trained with FullKD for one batch. 

\begin{table}[H]
\begin{center}
\begin{adjustbox}{max width=\linewidth}
\begin{tabular}{lS[table-format=2, round-mode=places, round-precision=0]S[table-format=2.1, round-mode=places, round-precision=1]}
\toprule
    \textbf{Method} & \textbf{$\Delta$ Angle $\downarrow$}  & \textbf{Norm Ratio}  \\
    \midrule
    Top-K 12 & \ang{58.0182578} &  2.4256 \\
    Top-K 50 & \ang{47.6145375} &  1.8133 \\ 
    Top-K 300 & \ang{29.7056098} &  1.2954 \\ 
    \rowcolor{gray!20}
    Random Sampling 12 & \bfseries \ang{3.94296081} & \bfseries 1.0027 \\

    \bottomrule
\end{tabular}
\end{adjustbox}
\end{center}
\caption{Comparing sparse KD gradients with FullKD}
\label{table: gradients}
\end{table}

We find that the gradients from using Random Sampling are extremely similar to the gradients obtained from FullKD -- with an angular difference of $\ang{4}$ and the same norm (cosine similarity of $0.998$, and relative error of $0.07$). Top-K methods on the other hand, have large angles and significantly different gradient norms even at $300$ tokens, compared to just $12$ unique tokens for Random Sampling.

\subsection{Variance and Bias of Sampling Methods}


While sampled distributions using Top-K have the least error for a single token, they inherently provide a biased estimate of the teacher distribution~(\cref{topk_error_proof}). This leads to the dissimilar gradients observed in \cref{sec:gradient}. While our method is always unbiased, it is also crucial for the estimator to exhibit low variance (error). Lower variance will result in better approximation of the teacher distribution and hence better gradient approximation.

For example, using $\tau=0$ in our proposal (sampling uniformly across the vocabulary) causes training to diverge, as the estimate is too noisy~(\cref{table: Random Sampling Temperature}). Similarly, using fewer tokens (with $\tau=1$) will have higher error -- but $12$ tokens seems to be sufficient~(\cref{table: Random Sampling Tokens}), and hence we use $12$ unique tokens in the rest of our experiments.

\subsection{Speed/Throughput Comparison}
In this section, we compare the speed in tokens/sec and TFlops for $300$M / $3$B student models with $3$B / $8$B teachers on $8$ H100 GPUs. Our (RS-KD) caching implementation is $1.7$ to $2.6$ times faster than FullKD, and only slightly slower ($\approx 10\%$) than CE. This overhead stems from computing the loss over the entire vocabulary for distillation compared to a single ground truth token for CE.

\begin{table}[H]
\begin{center}
\begin{adjustbox}{max width=0.9\linewidth}
\begin{tabular}{lcccc}
\toprule
   \textbf{} & \multicolumn{2}{|c|}{\textbf{Tokens/sec $\uparrow$}}   & \multicolumn{2}{c}{\textbf{TFlops $\uparrow$}}    \\
   \textbf{Method} & \textbf{300M}  & \textbf{3B} & \textbf{300M}  & \textbf{3B}   \\
    \midrule
    CE	        & $2.9$x	 &   $1.77$x & $330$	& 544 \\
    \rowcolor{gray!20}
    Random Sampling	    & $2.6$x     &	 $1.73$x & $295$	& 530 \\
    Full KD	    & $1.0$x	 &   $1.00$x & $100$	& 304 \\
    \bottomrule
\end{tabular}
\end{adjustbox}
\end{center}
\caption{Speed/Throughput Comparison.}
\label{table: speed}
\end{table}

\subsection{Storage Comparison}

For CE training, storing raw UTF-8 text for $100$B tokens requires $\approx0.5$TB storage for English (more for other languages). Storing tokenized data consumes $0.3$TB, assuming $3$ bytes per token. For FullKD storing the entire output distribution would requiring infeasible $10$PB of storage, assuming $1$byte for probability.

For sparse KD (KD) methods such as ours or Top-K, need to additionally store the Vocabulary Ids of the saved tokens. As detailed in \cref{sec:appendix_implmentation_concers}, we use $17$ bits for Vocabulary IDs, and $7$ bits for probabilities, totaling $24$ bits ($3$ bytes) per unique token. As we require only $12$ tokens~(\cref{table: Random Sampling Tokens}), we need only additional $3.6$TB of space, $25$x less than Top-300.


\begin{table}[H]
\begin{center}
\begin{adjustbox}{max width=0.95\linewidth}
\begin{tabular}{lS[table-format=6, round-mode=places, round-precision=0]S[table-format=1, round-mode=places, round-precision=0]S[table-format=5.1, round-mode=places, round-precision=1]}
\toprule
    \textbf{Method} & \textbf{Logits per}  & \textbf{Bytes per} & \textbf{Total} \\
    \textbf{} & \textbf{Train Token} & \textbf{Logit} & \textbf{Memory (TB)}  \\
    \midrule
    Full KD    & 100000	&  1	& 10000  \\
    Top-K 300 	& 300	& 3	& 90  \\
    \rowcolor{gray!20}
    Ours  & 12	& 3	& 3.6  \\
    Vanilla CE  & 1	& 3	& 0.3  \\
    \bottomrule
\end{tabular}
\end{adjustbox}
\end{center}
\caption{Storage Requirements for $100$B train tokens}
\label{table: Storage Comp}
\end{table}


\section{Results}
\label{sec:results}

\paragraph{Evaluation Tasks} We evaluate our method across multiple metrics -- LM loss on the pre-training dataset, Expected Calibration Error, the acceptance rate on speculative decoding of the teacher model, 0-shot NLU scores~(settings detailed in \cref{sec:appendix_nlu}, full scores in \cref{tab:all_downstream_results}) before and after Instruction Following training, and 0-shot NLG scores~(settings detailed in \cref{sec:appendix_IF}).

\subsection{Small-Scale Results}
\label{result:small-scale}

We train LLaMA-style $300$M student models using a $3$B teacher (hyper-parameters in \cref{hparams_300M}) for $10$B tokens, $1.5$x more than Chinchilla-optimal~\citep{chinchilla} number of tokens. Our proposed method achieves very similar performance and calibration compared to FullKD, while using only $12$ tokens~(\cref{table: Random Sampling Tokens}).

We also measure Speculative Decoding acceptance rate, as Top-$1$ agreement rate with the teacher has been shown to correlate with performance~\citep{DoesKnowledgeDistillation}. We find that our method again performs comparable to FullKD. 

Somewhat surprisingly, as the number of unique tokens is increased, random sampling achieves marginally better performance compared to FullKD. Prior work has found that perturbing teacher logits results in better KD~\citep{NotBlindlyImitate}, and we conjecture this sampling may achieve something similar.

\begin{table}[H]
\begin{center}
\begin{adjustbox}{max width=0.85\linewidth}
\begin{tabular}{S[table-format=2.1, round-mode=places, round-precision=1]S[table-format=1.2, round-mode=places, round-precision=2]ZS[table-format=.1, round-mode=places, round-precision=1]S[table-format=2.2, round-mode=places, round-precision=2]S[table-format=2.1, round-mode=places, round-precision=1]}
\toprule
   \textbf{Unique} & \textbf{LM}  & \textbf{\% CE to} & \textbf{ECE} & \textbf{Speculative} & \textbf{0-shot} \\
   \textbf{Tokens} & \textbf{Loss $\downarrow$}  & \textbf{FullKD $\uparrow$} & \textbf{\% $\downarrow$} & \textbf{Accept \% $\uparrow$} & \textbf{Score $\uparrow$} \\
    \midrule
    \text{CE}   & 2.80668 & 0\% & 0.38 & 59.94841 & 40.4  \\
    \midrule
    2.36 &	2.76939	&	62.46231156\%  & 1.036 & 61.4666 & 42.1  \\
    5.02 &	2.75299	&	89.93299832\% & 1.066 & 61.82802 & 42.6 \\
    \rowcolor{gray!20}
    12.05 &	2.74621	&	101.2897822\% & 0.783 & 61.84816 & 43.0  \\
    24.47	&	2.74606	&	101.5410385\% & 1.051 & 61.92517 & \bfseries 43.1 \\
    56.95	&	\bfseries 2.744	&	\bfseries 104.9916248\% & 0.902 & 61.96851 & 42.9 \\
    \midrule
    \text{FullKD} &  2.74698 & 100\% & \bfseries 0.73  &  \bfseries 62.01738 & 42.1\\
    \bottomrule
\end{tabular}
\end{adjustbox}
\end{center}\caption{Random Sampling KD ($3$B $\rightarrow$ $300$M)}
\label{table: Random Sampling Tokens}
\end{table}


\paragraph{Effect of Longer Training} On extending training of the student model for $100$B tokens ($16$x Chinchilla-optimal), our model again achieves performance comparable to FullKD, both in speculative decoding and in 0-shot NLU scores~(\cref{table: Random Sampling Long Internal}).

\begin{table}[H]
\begin{center}
\begin{adjustbox}{max width=0.85\linewidth}
\begin{tabular}{lS[table-format=1.2, round-mode=places, round-precision=2]S[table-format=1.1, round-mode=places, round-precision=1]S[table-format=2.1, round-mode=places, round-precision=1]S[table-format=2.1, round-mode=places, round-precision=1]}
\toprule
    \textbf{Method} & \textbf{LM} & \textbf{ECE} & \textbf{Speculative} & \textbf{0-shot} \\
    \textbf{} & \textbf{Loss $\downarrow$} & \textbf{\% $\downarrow$} & \textbf{Accept \% $\uparrow$} & \textbf{Score $\uparrow$} \\
    \midrule
    CE  & 2.48029 & 0.662 & 64.55876 & 45.0 \\
    \rowcolor{gray!20}
    Ours &	2.47658 & \bfseries 0.3 & 65.67473 & \bfseries 46.2 \\
    FullKD  &  2.48342 & 0.428 & \bfseries 65.76087 & \bfseries 46.2 \\
    \bottomrule
\end{tabular}
\end{adjustbox}
\end{center}
\caption{Random Sampling KD $100$B toks ($3$B$\rightarrow$$300$M)}
\label{table: Random Sampling Long Internal}
\end{table}

\subsection{Large-Scale Results}
\label{result:large-scale}

In order to replicate our findings with open-source LLMs on public datasets, we train student models using the LLaMA-$3$-$8$B model on the Fineweb-edu~\cite{finewebdatasetsdecantingweb} dataset. 

First, we train a $3$B LLaMA-style student using $100$B tokens (\cref{table: Random Sampling LLama}). The loss gap between Top-K KD and FullKD is much higher in this regime. On the contrary, the student trained using "Random Sampling KD" ($12$ unique tokens) achieves similar loss, calibration and speculative decoding acceptance rate with significantly better downstream and instruction following performance. The improvements observed in our small-scale experiments persist for larger models with much longer training.

\begin{table}[H]
\begin{center}
\begin{adjustbox}{max width=\linewidth}
\begin{tabular}{lS[table-format=1.3, round-mode=places, round-precision=2]S[table-format=1.1, round-mode=places, round-precision=1]S[table-format=2.1, round-mode=places, round-precision=1]S[table-format=2.1, round-mode=places, round-precision=1]S[table-format=2.1, round-mode=places, round-precision=1]}
\toprule
    \textbf{Method} & \textbf{LM} & \textbf{ECE} & \textbf{Speculative} & \textbf{0-shot}  & \textbf{IF SFT} \\
    \textbf{} & \textbf{Loss $\downarrow$} & \textbf{\% $\downarrow$} & \textbf{Accept \% $\uparrow$} & \textbf{Score $\uparrow$}  & \textbf{Score $\uparrow$} \\
    \midrule
    CE  & 2.37093 & 0.312 & 71.05489 & 55.6 & 54.5 \\
    \midrule
    Top-K 12 &	2.49963 & 4.664 & 72.97664 & 56.6 & 57.7 \\
    Top-K 50 &	2.3957 & 1.79 & 73.07028 & 57.1 & 58.3 \\
    \rowcolor{gray!20}
    Ours (12) &	2.35005 & \bfseries 0.195 & 73.18993 &  57.5 & \bfseries 59.4 \\
    Ours (12)+ & \bfseries 2.321 & 1.687 & \bfseries 73.45504 & \bfseries 57.9 & 59.1 \\
    \midrule
    FullKD  &  2.34446 & \bfseries 0.2 & 73.36377 & 57.5 & 58.4 \\
    \bottomrule
\end{tabular}
\end{adjustbox}
\end{center}
\caption{Comparing sparse KD methods, $8$B$\rightarrow3$B $100$B toks. The row `Ours (12)+' is described in \cref{sec:hard_easy}. }
\label{table: Random Sampling LLama}
\end{table}

\paragraph{Evaluation with LLM-as-a-judge on Generative Tasks}
\label{result:judge}

We also evaluate the $3$B models using Llama 3.1 $405$B Instruct~\cite{grattafiori2024llama3herdmodels} as a judge on five instruction following tasks. The student model trained with "Random Sampling KD" outperforms all other methods across all the evaluated tasks as seen in \cref{table: Random Sampling LLama Judge}.

\newcolumntype{G}{>{\columncolor{gray!15}}S[table-format=2.1]}

\begin{table}[H]
\begin{center}
\begin{adjustbox}{max width=0.95\linewidth}
\begin{tabular}{lS[table-format=2.1]S[table-format=2.1]S[table-format=2.1]GS[table-format=2.1]}
\toprule
        \textbf{Dataset} & \textbf{CE} & \textbf{Top-K} & \textbf{Top-K} & \textbf{Ours}  & \textbf{FullKD} \\
        \textbf{} & \textbf{} & \textbf{12} & \textbf{50} & \textbf{12}  & \textbf{} \\
    \midrule
Dolly & 64.2 & 59.0 & 65.4 & \bfseries 71.3 & 66.1 \\
SelfInst & 64.6 & 60.9 & 63.4 & \bfseries 73.1 & 66.1 \\
Vicuna & 49.1 & 48.9 & 53.1 & \bfseries 58.2 & 56.9 \\
S-NI & 62.4 & 63.4 & 62.6 & \bfseries 63.8 & 60.7 \\
UnNI & 60.4 & 58.0 & 58.3 & \bfseries 61.4 & 61.0 \\
\midrule
Avg & 60.2 & 58.0 & 60.6 & \bfseries 65.6 & 62.2 \\
    \bottomrule
\end{tabular}
\end{adjustbox}
\end{center}
\caption{Evaluations of $3$B models on downstream generative tasks, with LLM-as-judge ($8$B $\rightarrow$ $3$B)}
\label{table: Random Sampling LLama Judge}
\end{table}

\paragraph{Effect of Student Size}

\begin{figure}[H]
     \centering
     \includegraphics[width=0.75\linewidth]{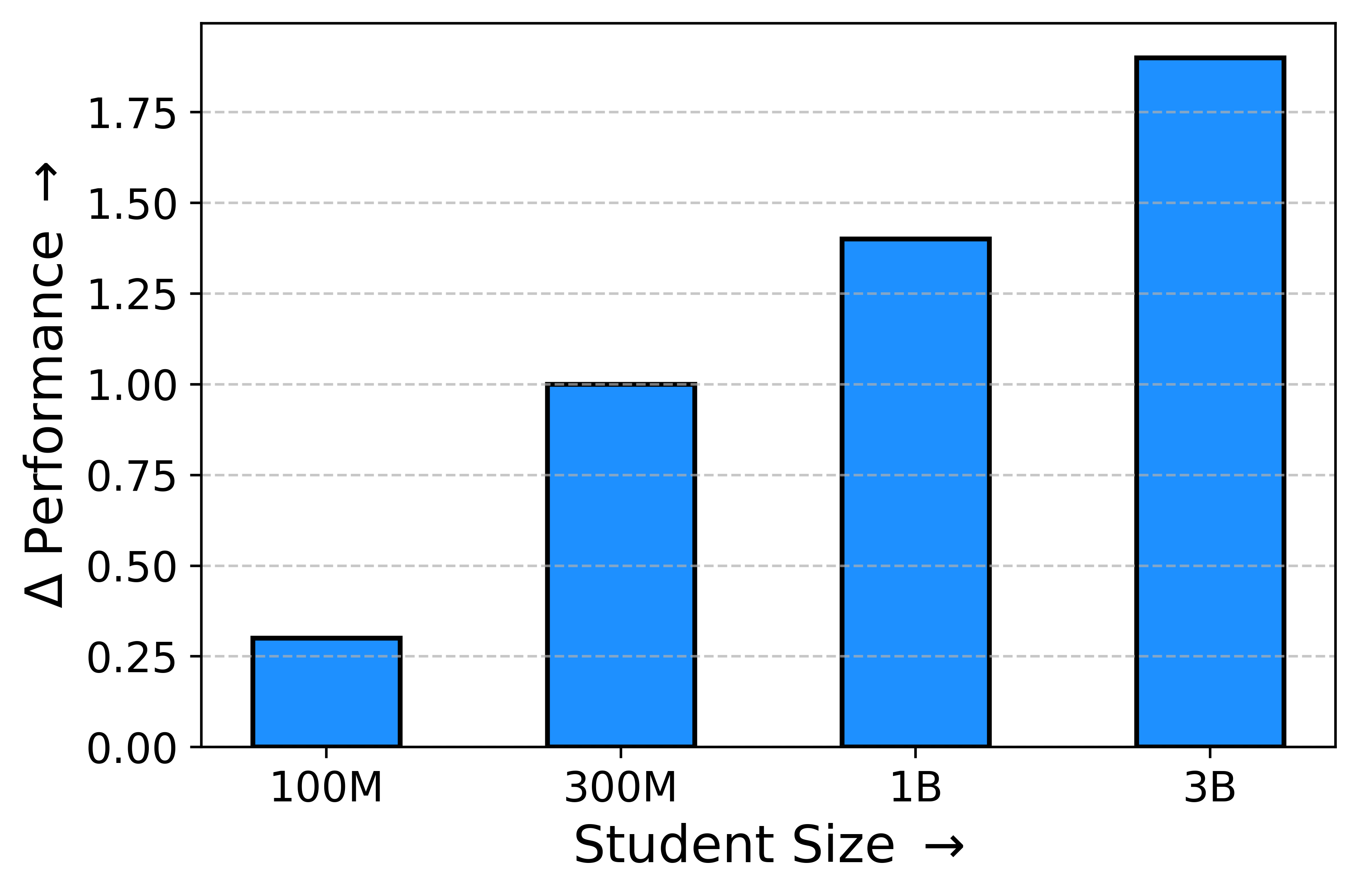}
     \caption{Downstream Improvements vs Student Size}
     \label{fig:diff_vs_student_size}
 \end{figure}

We also vary the student sizes, training $100$M, $300$M, $1$B and $3$B all trained using LLaMA-$3$-$8$B as teacher, for $30$x model-size tokens. The average performance on $0$-shot downstream evaluations using "Random Sampling KD" over CE consistently improves as the student model size increases~(\cref{fig:diff_vs_student_size}). While similar increasing trends have been previously observed for Top-K pre-training in \citet{PretrainingDistillationLarge}, they report a \textit{fall} in performance for smaller student models. We conjecture that this may be attributed to the issues with Top-K KD we highlight in this work.

\subsection{Orthogonal Improvements to KD}
\label{sec:hard_easy}

Some orthogonal methods have also been proposed in the literature to improve the performance of FullKD. In this section, we show that these approaches can also be applied to "Random Sampling KD". Adding a combination of KLD and CE losses is often used during training~\citep{AppleIntelligenceFoundation, PretrainingDistillationLarge, DualSpaceKnowledgeDistillation} where the final loss is defined as $L = \alpha \cdot L_{CE} + (1-\alpha) \cdot L_{KLD}$ where $\alpha$ is the CE weight. Some prior works~\citep{RevisitingKnowledgeDistillation, zhao2022decoupled, LionAdversarialDistillation, PerformanceGuidedLLMKnowledge} use different training modes for different tokens based on teacher's confidence/score in the target, where higher a score indicates that a token is easy to learn. 

\paragraph{Setup} We apply a similar adaptive method to our "Random Sampling KD" by categorizing tokens in a batch as ``Easy'' and ``Hard'' based on their target confidence percentile. Hard tokens use a higher learning rate (by a factor of ``LR Ratio'') compared to the easy tokens during training, while the average LR is kept constant. We train $300$M models using a $3$B teacher, and simultaneously vary the CE weight and the LR ratio together and report the '\% CE to FullKD' metric.

\paragraph{Results} As seen in \cref{table: Adaptive LR}, these methods enable "Random Sampling KD" to surpass FullKD. The best model is achieved with $0.1$ CE weight and $2.0$ LR Ratio. We further apply this approach to train a $3$B student with $8$B teacher for $100$B tokens. This model (the row Ours (12)+ in \cref{table: Random Sampling LLama}), further improves on "Random Sampling KD" in LM loss, speculative acceptance rate, and 0-shot NLU scores.

\paragraph{Caveats} However, this model does not improve as much after Instruction Tuning. We conjecture that up-weighing the "Hard" examples in the LR tends to effectively up-weigh the tail of the distribution. This was evidenced by the relatively higher calibration error of this model -- we find that this model is \textit{under-confident} in its predictions. While this improves the pre-training scores, it negatively impacts downstream fine-tuning of this model.

\begin{table}[H]
\begin{center}
\begin{adjustbox}{max width=0.7\linewidth}
\begin{tabular}{lS[table-format=3, round-mode=places, round-precision=0]S[table-format=3, round-mode=places, round-precision=0]S[table-format=3, round-mode=places, round-precision=0]S[table-format=3, round-mode=places, round-precision=0]}
\toprule
    \textbf{LR Ratio} & \multicolumn{4}{c}{\textbf{CE Weight $\mathbf{\alpha}$}} \\
    \cmidrule{2-5}
     & \textbf{0.3} & \textbf{0.2} & \textbf{0.1} & \textbf{0.0} \\
    \midrule
    $\mathbf{1.0}$ & 100.8203926 & 111.3096982 & 94.55024905 & 98.00761793 \\
    $\mathbf{1.5}$ &	123.7620861 & 120.6856138 & 119.6015236 & 110.8995019 \\
    $\mathbf{2.0}$  &  115.6167594 & 123.8499854 & \bfseries 124.5238793 & 112.4523879 \\
    \bottomrule
\end{tabular}
\end{adjustbox}
\end{center}
\caption{ `\% CE to FullKD' with Orthogonal Improvements to Random Sampling KD ($8$B $\rightarrow$ $300$M)}
\label{table: Adaptive LR}
\end{table}

\subsection{Comparison with Prior Works}

In \cref{table: Prior Work}, we compare our sampling approach with those from prior works. For \citet{DistillationTokensAre}, we use Top-$50$, and for \citet{PretrainingDistillationLarge}, Top-$100$ with $p=0.98$. We also recreate these works including other sampling-orthogonal changes. \citet{DistillationTokensAre} uses a different LR for harder tokens, and adds the CE Loss to training. For \citet{PretrainingDistillationLarge}, we implement the temperature before softmax, and the WSD scheduler for the relative weight of CE and KD. Our method significantly outperforms these prior works.

\begin{table}[H]
\begin{center}
\begin{adjustbox}{max width=\linewidth}
\begin{tabular}{lS[table-format=1.2, round-mode=places, round-precision=2]S[table-format=-3\%, round-mode=places, round-precision=0]S[table-format=1.1, round-mode=places, round-precision=1]S[table-format=2.1, round-mode=places, round-precision=1]}
\toprule
    \textbf{Method} & \textbf{LM}  & \textbf{\%CE} & \textbf{ECE} & \textbf{Spec.} \\
    \textbf{} & \textbf{Loss $\downarrow$} & \textbf{to FullKD $\uparrow$} & \textbf{\% $\downarrow$} & \textbf{Accept \% $\uparrow$}  \\
    \midrule
    CE & 2.80668 & 0\% & 1.154 & 59.95 \\
    \midrule
    \citet{PretrainingDistillationLarge}* 	& 2.78	& -47\%	& 1.7 &	61.85 \\
    \citet{PretrainingDistillationLarge} 	& 2.85	& -78\%	& 1.4 &	61.49 \\
    \citet{DistillationTokensAre}*    & 2.80	&  5\%	& 2.2 &	61.89 \\
    \citet{DistillationTokensAre}    & 2.77	&  57\%	& 1.9 &	60.99 \\
    \rowcolor{gray!20}
    Ours  & \bfseries 2.74	& \bfseries 100\%	& 0.9 &	\bfseries 61.97 \\
    \midrule
    FullKD  &  2.74698 & \bfseries 100\% & \bfseries 0.73 & \bfseries 62.02 \\
    \bottomrule
\end{tabular}
\end{adjustbox}
\end{center}
\caption{Comparison with Prior Works. Rows marked with * only use the sampling method. ($3$B $\rightarrow$ $300$M)}
\label{table: Prior Work}
\end{table}

\section{Ablations}

\subsection{Proposal Distributions}

Choosing the optimal sampling temperature $t$ can reduce the variance of the probability estimates, by allowing a trade-off between sampling more varied tokens, vs. obtaining more accurate estimates for higher-probability tokens. While this optimal temperature would depend on the exact shape of the distribution (and hence the teacher model), numerical simulations show that $t \in [0.8,1.2]$ results in the lowest variance. The post-training performance of these was also comparable~(\cref{table: Random Sampling Temperature}). 

While a better proposal distribution may be obtained following Optimal Experimental Design~\citep{TheoryOptimalExperiments}, our sampling method performs comparable to FullKD, hence for simplicity we choose proposal with $t=1.0$ in this work.

\begin{table}[H]
\begin{center}
\begin{adjustbox}{max width=\linewidth}
\begin{tabular}{S[table-format=1.1]S[table-format=2, round-mode=places, round-precision=0]ZS[table-format=1.2, round-mode=places, round-precision=2]S[table-format=.1, round-mode=places, round-precision=1]S[table-format=2.1, round-mode=places, round-precision=1]S[table-format=2.1, round-mode=places, round-precision=1]}
\toprule
    \textbf{Sample} & \textbf{Unique} & \textbf{Noise \%} & \textbf{LM} & \textbf{ECE}  & \textbf{0-shot} & \textbf{Speculative} \\
    \textbf{Temp} & \textbf{Tokens} & \textbf{Single} & \textbf{Loss $\downarrow$} & \textbf{\% $\downarrow$} & \textbf{Score $\uparrow$} & \textbf{Accept \% $\uparrow$} \\
    \midrule
    0.0 & 57	& 99 &  $\infty$ & \text{-} & \text{-} & \text{-} \\
    0.8 & 56.939	& 24.82 & \bfseries	2.74034	 & \bfseries 0.700 & 42.4  & 61.89992 \\
    \rowcolor{gray!20}
    1.0 & 53.78	& 19.99 &	2.74621	 & 0.783 & \bfseries 43 & \bfseries 61.84816 \\
    1.2 & 57.44	& 16.80 &	\bfseries 2.73956	 & 0.794 & 42.2 & 61.89717 \\
    \bottomrule
\end{tabular}
\end{adjustbox}
\end{center}
\caption{Proposal Temperature Ablation ($3$B$\rightarrow$$300$M)}
\label{table: Random Sampling Temperature}
\end{table}


\subsection{Effect of Adapting Teacher}

\citet{LLMPruningDistillation} found that if the student is being trained on a data distribution different from the teacher's pre-training data, the teacher should first be adapted (finetuned) on this data by training for a short while. We also observe the same -- when training a $300$M student on Fineweb-edu data with the LLaMA-$3$-$8$B model as teacher, using the original teacher model directly yields only a small improvement over CE~(\cref{table: Adapting Teacher}). After teacher adaptation for $50$B tokens, this increases significantly.

\begin{table}[H]
\begin{center}
\begin{adjustbox}{max width=0.8\linewidth}
\begin{tabular}{lS[table-format=1.2, round-mode=places, round-precision=2]S[table-format=2.1, round-mode=places, round-precision=1]}
\toprule
    \textbf{Method} & \textbf{LM Loss $\downarrow$}  & \textbf{0-shot Score $\uparrow$}  \\
    \midrule
    CE  & 2.99378 & 40.1 \\
    KD w/o adapt  & 2.9821 & 40.2 \\
    \rowcolor{gray!20}
    KD w adapt  & \bfseries 2.95965 & \bfseries 41.1 \\
    \bottomrule
\end{tabular}
\end{adjustbox}
\end{center}
\caption{Adapting Teacher Model on Pre-training Dataset ($8$B $\rightarrow$ $300$M)}
\label{table: Adapting Teacher}
\end{table}

\subsection{Effect of Different Student Architecture}

Our method is independent of the model architecture, and is equally applicable to other models such as Qwen~\citep{qwen2.5}. Using the above LLama-$3$-$8$B as teacher, we train a $0.5$B Qwen-style model (same architecture as Qwen$2.5$-$0.5$B) using our Random Sampling Method and with vanilla CE for $10$B training tokens. Our method improves over CE as shown in \cref{table: Qwen}.

\begin{table}[H]
\begin{center}
\begin{adjustbox}{max width=0.65\linewidth}
\begin{tabular}{lS[table-format=1.2, round-mode=places, round-precision=2]S[table-format=2.1\%, round-mode=places, round-precision=1]}
\toprule
    \textbf{Method} & \textbf{LM}  & \textbf{Speculative} \\
    \textbf{} & \textbf{Loss $\downarrow$} & \textbf{Accept \% $\uparrow$}  \\
    \midrule
    CE	& 2.99 &	58.9\% \\
    \rowcolor{gray!20}
    Ours	& \bfseries 2.95 & \bfseries	60.0\% \\
    \bottomrule
\end{tabular}
\end{adjustbox}
\end{center}
\caption{Pre-training Qwen-style models ($3$B $\rightarrow$ $0.5$B)}
\label{table: Qwen}
\end{table}

\subsection{Choice of Loss/Divergence Function}

We also experiment with alternative loss/divergence functions, by training $300$M students with $8$B Llama-$3$ teacher for $10$B tokens. Some prior works~\citep{ComparingKullbackLeiblerDivergence, RethinkingKullbackLeiblerDivergence, MiniLLMKnowledgeDistillation, DistiLLMStreamlinedDistillation} find alternative objectives such as Reverse KL Divergence, Mean Squared Error as superior, while other works~\citep{KnowledgeDistillationApprox, FDivergenceMinimizationSequenceLevel, CompactLanguageModels, PretrainingDistillationLarge} have observed the opposite. In \cref{table: Loss Ablation}, we observe that vanilla forward KLD outperforms other objectives.

\begin{table}[H]
\begin{center}
\begin{adjustbox}{max width=\linewidth}
\begin{tabular}{lcccccc}
\toprule
    \textbf{Metric} & \textbf{CE} & \textbf{L1} & \textbf{MSE} & \multicolumn{3}{c}{\centering \textbf{KLD}}   \\ 
    \cmidrule{5-7}
     & \textbf{} &  & & \textbf{R} & \textbf{F+R} & \textbf{F}  \\
    \midrule
    LM Loss $\downarrow$ & $2.81$ & $\infty$ & $5.38$ & $3.37$ & $2.78$  & $\mathbf{2.75}$  \\
    \bottomrule
\end{tabular}
\end{adjustbox}
\end{center}
\caption{Loss Ablation. F and R in KLD refer to forward and reverse KLD respectively.}
\label{table: Loss Ablation}
\end{table}

\section{Related Work}

Knowledge Distillation ~\citep{DistillingKnowledgeNeural} has often been used to improve smaller LLMs \citep{jiao2019tinybert, sanh2020distilbertdistilledversionbert, LLMPruningDistillation, CompactLanguageModels, MiniLMv2MultiHeadSelfAttention, MiniLLMKnowledgeDistillation, CrossTokenizerDistillationUniversal}. Many works focus on using teacher models for dataset generation/filtering~\citep{SequenceLevelKnowledgeDistillation, NotBlindlyImitate, FDivergenceMinimizationSequenceLevel, TextbooksAreAll, LionAdversarialDistillation, MiniPLMKnowledgeDistillation, PerformanceGuidedLLMKnowledge}. These methods are somewhat complementary to our method -- our work is agnostic to the source of the pre-training data corpus, and focuses on distilling the teacher model's logits on this data.

Similar to our work, \citet{FIRSTTeachReliable} stores the Top-$5$ teacher probabilities from an LLM for training smaller students. They also observe that distillation with Top-K tokens leads to over-confident students -- which they solve  by employing temperature scaling. By sampling from the teacher distribution, our method offers a principled approach of achieving a calibrated student~(\cref{fig:ece-compare}). While they observe mis-calibration of their teacher as well, pre-trained LLMs are well-calibrated, but alignment may degrade this calibration~\citep{CalibrationLargeLanguage, UnderstandingCalibrationTransfer}. We find both our $3$B as well as Llama $8$B teachers well calibrated, as they are not instruction-tuned models.

Closest to our work are \citet{DistillationTokensAre}, \citet{PretrainingDistillationLarge} and \citet{gemma3}. \citet{DistillationTokensAre} also observe that distillation improves student model performance -- but they store Top-$5\%$ of the teacher logits, which is prohibitively large for modern LLMs ($6400$ for the Llama$3$ model) -- we successfully bring this down to $12$ logits in this work.

\citet{PretrainingDistillationLarge} explores caching teacher logits in Knowledge Distillation in pre-training of LLMs utilizing Top-K with Top-p. They also conclude that forward KLD outperforms other objectives, adding CE loss improves distillation, and increasing performance improvement on scaling the model size and pre-training corpus. However, they observe a \textit{fall} in performance on smaller students -- vanilla Top-K may reduce model performance if $K$ is not large enough as we show in \cref{table: Vanilla Top-K}. Our method remedies this issue, matching FullKD with significantly sparser tokens.

Contemporaneous work Gemma3~\citep{gemma3} also used Knowledge Distillation for pre-training. Their method seems to be the same as our approach, sampling teacher logits weighed by original teacher probabilities, using cross-entropy loss on the sampled tokens. They successfully apply this method for training model up-to $27$B params for $14$T tokens, showing that our method can scale to very large models and tokens.

\section{Conclusion}

In this work, we identified key issues of bias and tail supervision with sparse teacher logits for Knowledge Distillation. We theoretically proved and empirically verified these claims in both synthetic and real-world scenarios, and proposed an importance-sampling based method to rectify them. By preserving gradients and logits distribution in expectation, we enable significantly sparser logits than prior methods. Our method maintains model performance while utilizing only $0.01\%$ of pre-computed teacher logits, across a range of model sizes, training tokens, and evaluation metrics.

\section*{Limitations}

Due to limited compute resources, we were only able to experiment upto $3$B scale models trained for $100$B tokens. Training longer with larger models should be explored, but our experiments indicate the benefits of our model only increase with model scale. Representation matching, which distills intermediate activations from the teacher, may improve distillation further. However, caching teacher representations due to limited compute resources was a primary requirement for this work, which rendered representation matching infeasible. Lastly, more sophisticated sampling schemes can also be explored, but we did not attempt that as our methods already achieved the desired outcome of matching full KD with low storage requirements. 


\nocite{*}

\bibliography{custom}

\appendix
\label{appendix}

\section{Proofs}
\label{sec:appendix_proofs}

\subsection{Backward Gradient through Softmax-KL Divergence Loss}
\label{sec:sfotmax-kld-grad}
The output probability $\mathbf{p}$ is defined in terms of the model's logits $\mathbf{x}$
\begin{align*}
    \mathbf{p} &= \mathrm{Softmax}(\mathbf{x})\\
    p_i &= \frac{e^{x_i}}{\sum_{j=1}^{|V|}e^{x_j}}\\
\end{align*}

The gradient through Softmax \cite{iwana2019explaining} is:
\begin{align*}
    \frac{\partial p_i}{\partial x_j} &=  p_i.(1\{i=j\} - p_j) \\
\end{align*}
Given a target probability distribution $\mathbf{t}$, the KL divergence loss is defined as:
\begin{align}
    L =  \sum_{i=1}^{|V|} t_i \log \frac{t_i}{p_i}  \label{eq:kld_loss}
\end{align}

For Softmax-KL Divergence Loss, the gradient flowing to the $j_{th}$ logit $x_{j}$ can be calculated  as follows:

\begin{align*}
    \frac{\partial L}{\partial x_j} &= - \sum_{i=1}^{|V|} t_i \frac{1}{p_i} \frac{\partial p_i}{\partial x_j} \\
    &= \sum_{i=1}^{|V|} t_i.(p_j - 1\{i=j\}) \\
    &= (\sum_{i=1}^{|V|} t_i).p_j - t_j
\end{align*}

If the full teacher distribution is provided $\sum_{i=1}^{|V|} t_i =1$. However, in the most generalized form, the gradient through Softmax-KL divergence loss can be written as: 
\begin{align}
    \frac{\partial L}{\partial x_j} &= (\sum_{i=1}^{|V|} t_i).p_j - t_j \label{eq:loss_grad_general}
\end{align}

\subsection{Cross Entropy Loss}
The cross entropy loss $L$ defined as follows:
\begin{align*}
    L_{CE} &= - \sum_{i=1}^{|V|} t_i \log p_i  \\
    & = L_{KLD} - \sum_{i=1}^{|V|} t_i \log t_i 
\end{align*}
Compared to the KLD loss, the additional term ($\sum_{i=1}^{|V|} t_i log t_i$) is independent of the student model. Hence, the gradient for CE loss remains the same as that computed for KL Divergence loss in \cref{eq:kld_loss}. For cross entropy (and similarly for FullKD with KLD loss), $\sum_{i=1}^{|V|} t_i = 1$. Hence, the gradient can be further simplified to:
\begin{align*}
    \frac{\partial L}{\partial x_j} &=  {p_{_j}} - t_j
\end{align*}
In this case, the theoretical optima lies at the point where the predicted probabilities $\mathbf{p}$ become same as target probabilities $\mathbf{t}$ across the vocabulary, resulting in $0$ gradient and minimum loss. 

\subsection{Vanilla Top-K has the Least L1 Error, but is a Biased Estimate}
\label{topk_error_proof}

For a given distribution $\mathbf{t}$, if only $K$ probabilities from $\mathbf{t}$ must be kept, and they are then normalized to sum to $1$, we show that selecting the Top $K$ probabilities results in the least $L_1$ error.

Let $\mathbf{K}$ be the set of tokens selected. Let $a=\sum_{j\in K} t_j$. This can be viewed as constructing a new distribution $\mathbf{v}$, where normalizing the probabilities
\begin{align*}
v_i&=\frac{t_i}{a}, i \in K, \\
v_i&=0, i \not\in K
\end{align*}
Then the $L_1$ error between $\mathbf{t}$ and $\mathbf{v}$ is
\begin{align*}
    L_1 &= \sum_{i} | t_i - v_i| \\
    &= \sum_{i\in K} | t_i - t_i/a| + \sum_{i\not\in K} | t_i - 0| \\
    &= (1/a -1)*\sum_{i\in K}  t_i  + (1- \sum_{i\in K}  t_i) \\
    &= (1/a -1)*a   + (1- a) \\
    &= 2*(1-a) \\
\end{align*}

Hence $L_1$ will be minimized when $a$ is the largest, which will happen when the $K$ largest probabilities are selected.

However, note that this gives us a biased estimate, as $E[v_i] = 0 \neq E[t_i], i \not\in K$.

\subsection{Vanilla Top-K KD provides scaled teacher as target}
\label{naive_upscales}
We can restrict the target probability to a subset of tokens in our vocabulary. If we select $\mathbf{K}$ as the set of tokens with top-k probabilities, then the loss is defined as follows:
\begin{align*}
    L =  \sum_{i\in K} t_i \log \frac{t_i}{p_i}  
\end{align*}

This can be viewed as zeroing out the non-top-k target probabilities in the original KLD loss. In this case, the gradient flowing to the logits are (\cref{eq:loss_grad_general}):

\begin{align}
    \frac{\partial L}{\partial x_j} &= (\sum_{i\in K} t_i). {p_{_j}} - t_j \label{eq:loss_topk}
\end{align}

If $j \notin K $, the gradient is $(\sum_{i\in K} t_i). {p_{_j}} $. As opposed to the previous case, model's optima lies at the point where non-top-k probabilities are 0 and hence the student is \textbf{under-confident} in the non-top-k probabilities. Similarly, the top-k predicted probabilities $\mathbf{p}$ are a scaled up version of the target probabilities $\mathbf{t}$ across the top-k tokens, $p_i = \frac{t_i}{(\sum_{j\in K} t_j)}$, hence making the student \textbf{overconfident} in top-k probability predictions.
At this optima, the gradient is 0 (but the loss is negative).

\subsection{Ghost Token Backward}
\label{ghost_token}
One possible solution to the above discussed problem is to add a ghost token which accounts for the remainder of the probability. This ghost token ensures that the sum of probability outside the top-k region is exactly the same for the teacher and student. Ideally, it would ensure that the top-k tokens receive the exact teacher probability as the target. The modified loss function is written below-
\begin{align*}
    L = \Bigg( \sum_{i\in K} t_i & \log \frac{t_i}{p_i} + \\ & (1-\sum_{i\in K} t_i)log \Big(\frac{1-\sum_{i\in K} t_i}{1-\sum_{i\in K} p_i}\Big) \Bigg) 
\end{align*}

Let us consider the second term in the loss and find its gradient
\begin{align*}
    L_{ghost} &= (1-\sum_{i\in K} t_i)log \Big(\frac{1-\sum_{i\in K} t_i}{1-\sum_{i\in K} p_i}\Big) \\
    \frac{\partial L_{ghost}}{\partial x_j} &= \Big(\frac{1-\sum_{i\in K} t_i}{1-\sum_{i\in K} p_i}\Big) . \sum_{i=1}^{k} \frac{\partial p_i}{\partial x_j} \\
    = \Big(&\frac{1-\sum_{i\in K} t_i}{1-\sum_{i\in K} p_i}\Big) . \sum_{i=1}^{k} p_i.(1\{i=j\} - p_j)
\end{align*}

The gradient becomes:
\begin{align*}
\frac{\partial L_{\text{ghost}}}{\partial x_j} = \begin{cases}
\left(1 - \sum_{i\in K} t_i\right) p_j  & j  \in K, \\
-\Big(\frac{1-\sum_{i\in K} t_i}{1-\sum_{i\in K} p_i}\Big) p_j \sum_{i\in K} p_i & \text{else}.
\end{cases}
\end{align*}

Next we can add the gradient from top-k KD loss \cref{eq:loss_topk} and ghost token loss to obtain the final gradient
\begin{align*}
\frac{\partial L}{\partial x_j} = \begin{cases}
\left(p_j - t_j\right)  & j \in K, \\
\Big(\frac{\sum_{i\in K} (t_i - p_i)}{1-\sum_{i\in K} p_i}\Big) p_j & \text{else}.
\end{cases}
\end{align*}

For the non top-k tokens, the gradients can be rewritten as

\begin{align*}
\frac{\partial L_{\text{ghost}}}{\partial x_j} = p_j - \Big(\frac{1 - \sum_{i\in K} t_i }{1-\sum_{i\in K} p_i}\Big) p_j \notin K
\end{align*}

By adding the ghost token, the top-k tokens get the same gradient as KLD loss with FullKD, while the remaining tokens receive gradient in proportion of their predicted probability $p_i$. The target probability for non top-k tokens is $\Big(\frac{1 - \sum_{i\in K} t_i }{1-\sum_{i\in K} p_i}\Big) p_j$. In this case, if the predicted probability distribution is exactly the same as that of teacher probability only for top-k tokens, the gradient becomes 0 and loss becomes minimum.

\subsection{Random Sampling KD provides Unbiased Estimates}
\label{sec:RS_unbiased}

Our method Random Sampling KD uses importance sampling. By definition, importance sampling estimator is an unbiased estimator~\citep{AdvancesImportanceSampling}. We provide a short intuition of this below for temperature $t=1$.

We sample token ids $N$ times with replacement from proposal distribution $q_i=p_i$.

Each occurrence is assigned a likelihood ratio $\frac{p_i}{q_i}=1$, and then normalized by dividing by $N$.

The expected counts of token $i$ will then be $\frac{q_i*N}{N}=q_i=p_i$. Hence this sampling is unbiased.

\subsection{Unbiased Sampling preserves gradients in expectation}
\label{sec:grads_preserved}

For any partial knowledge distillation scheme which sub-samples the full distribution, the expected gradients at the logits will be preserved in expectation if sampling is unbiased.

\textbf{Proof:} The gradient $g_j$ for the logit $x_j$ through the softmax-KL divergence loss is (replacing $\sum_{i=1}^{|V|} t_i = 1 $ in \cref{eq:loss_grad_general}))

\begin{align}
    g_j &= p_j - t_j     
\end{align}

Taking expectations on both sides
\begin{align*}
    E[g_j] &= E[p_j] - E[t_j]     
\end{align*}

Similarly, for a sub-sampling method which reduced $\mathbf{t} \rightarrow\mathbf{t^s}$, expected gradient is as follows
\begin{align*}
    E[g^s_j] &= E[p_j] - E[t^s_j]     
\end{align*}

The gradients at the logits are preserved in expectation if $E[t_j]  = E[t^s_j]  $ and the sub-sampling process is unbiased. 

\section{Synthetic Examples}
\label{sec:synthetic}

\paragraph{Visualizing Target Probabilities} We generate a Zipf distribution where the probability of $i^{th}$ token is proportional $\dfrac{1}{i}$. Next we select tokens and assign them probabilities based on different sparse knowledge distillation methods. We plot these probabilities with the ground truth FullKD probabilities to visualize the alignment of sparse KD target distributions with FullKD. 


\paragraph{Calibration on Synthetic Classes} As discussed in the main paper and the psuedocode (\cref{toy_example_pseduo}), we generate synthetic data by generating random points around randomly chosen class means with Gaussian error distribution. We use a simple $3$-layer MLP as our model. We train the model using different sparse KD techniques and FullKD and plot the mean accuracy after binning the probabilities. 

\paragraph{Calibration on CIFAR-100} We follow the exact same methodology as the synthetic classification while using CIFAR-100 task and a weaker/smaller version of ResNet-18 model.

\section{Number of Sampling Rounds for Given Number of Effective Tokens}
\label{sec:sampled_vs_unique}
\begin{figure}[H]
     \centering
     \includegraphics[width=\linewidth]{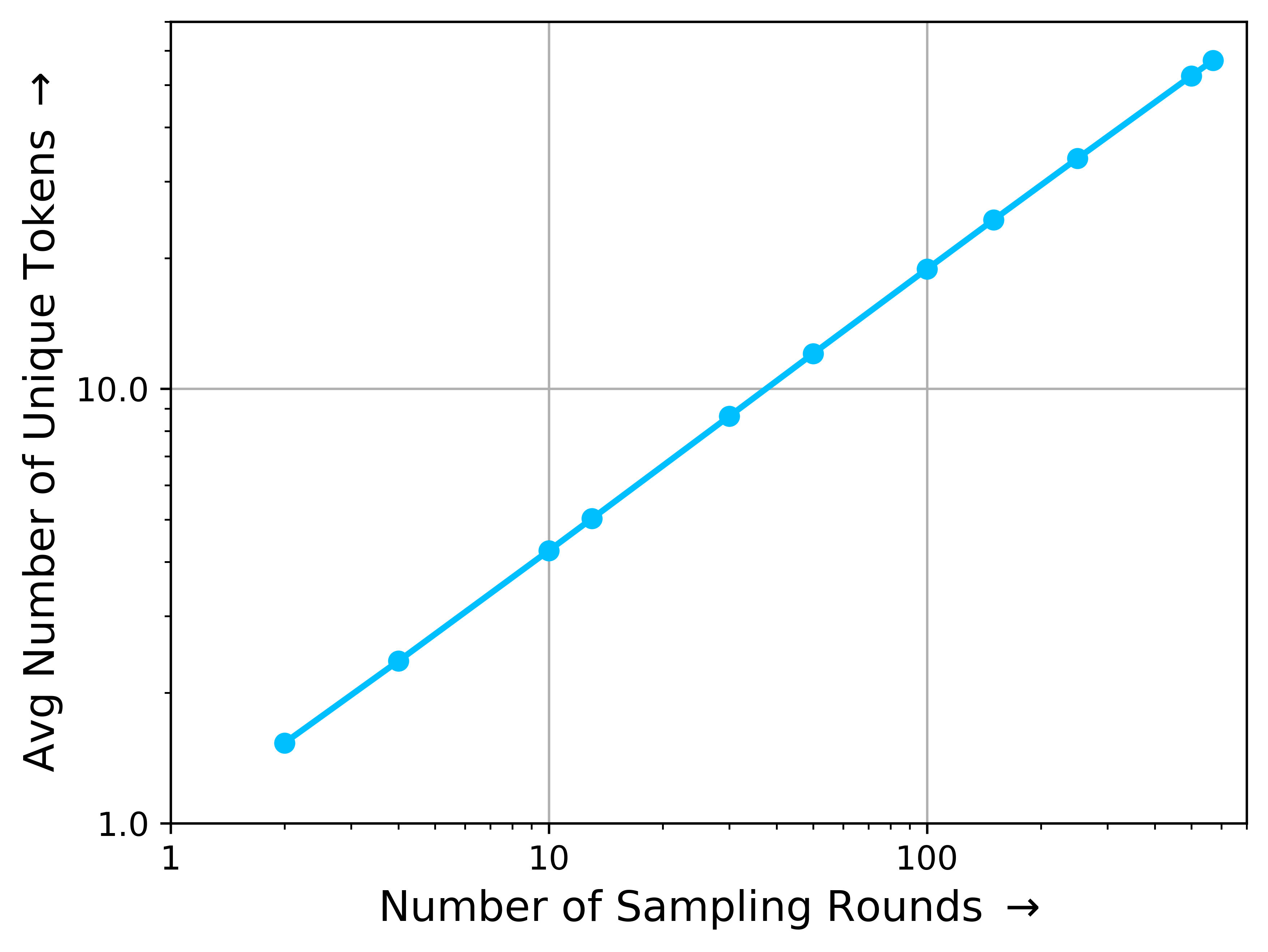}
     \caption{Number of unique tokens sampled vs sampling rounds}
     \label{fig:unique_tokens}
 \end{figure}

For a fair comparison between Top-$K$ KD and random sampling methods, the number of sampling rounds $N$ were chosen such that the number of unique tokens sampled match $K$. This will be specific to the dataset and the teacher model. For example, $N=50$, we find $K=12$. The relationship between the two for pre-training data is shown in \cref{fig:unique_tokens} (log-log scale), and is almost perfectly linear, showing an approximate power-law relationship.

\section{Implementation Concerns}
\label{sec:appendix_implmentation_concers}
\subsection{Quantization for Teacher Probabilities}

For our vocab size $V=100000$, our token ids require $log_2(V)=17$ bits. We store the byte-aligned data, which leaves us with $24-17=7$ bits for teacher probabilities. As probabilities are in range $0..1$, for Top-$K$ method, we use the $7$ bits to split the $0$ to $1$ range into $2^7$ equal intervals. This resulted in slightly lower performance compared to storing the probabilities in fp16. Instead switching to ratio encoding with sorted Top-$K$ probabilities resulted in significantly reduced quantization error to almost $0$, and results matched that of using un-quantized probabilities.

In the case of our proposed random sampling, we use $50$ sampling rounds, so token probabilities can only be of the form $x/50$, where $x$ is some integer. As this is less than $2^7$, we can store all of these exactly in $7$ bits by only storing the numerator. If sampling rounds are increased beyond $128$, ratio encoding with sorted probabilities can be used instead.

\subsection{Efficiency Concerns}

Naively implementing the sampling and the loss calculation incurred significant memory usage, due to the large vocabulary size. Manual backward and forward for the softmax KLD needed to implemented (via plain Pytorch, custom kernels were not created). Writing and reading the logits needed to be streamlines via shared memory ring buffers and async writer processes, so as to not block the GPU.

\subsection{Aligning Teacher and Student Sequences}

In our pre-training, we pack shuffled training documents to maximum sequence length, but we do not mask attention across document boundaries due to efficiency reasons. In our initial implementation, different shuffling seed was used between the teacher (during inference)and student (during training) -- This resulted in the prefix-context of tokens seen by the teacher and student not being aligned after the first document boundary. This had a surprisingly large effect on student model performance, particularly if smaller sequence lengths were used during teacher inference. We conjecture that with longer sequence lengths, far-away tokens from other documents will have less of an impact on the distribution of the logits. After fully aligning the teacher and student sequences, this effect was eliminated, and the offline run was within random error of the online run.

\begin{table}[ht]
\begin{center}
\begin{adjustbox}{max width=\linewidth}
\begin{tabular}{lS[table-format=4]S[table-format=1.3, round-mode=places, round-precision=3]S[table-format=3\%, round-mode=places, round-precision=0]}
\toprule
    \textbf{Shuffle Seeds} & \textbf{Seq Len}  & \textbf{LM Loss} & \textbf{\% CE to online}  \\
    \midrule
    Different & 1024 & 2.75958 & 78.89447236 \\
    Different & 4096 & 2.75303 & 89.86599665 \\
    Same & 4096 & \bfseries 2.74933 & \bfseries 96.06365159 \\
    \bottomrule
\end{tabular}
\end{adjustbox}
\end{center}
\caption{Effect of aligning teacher and student sequences, with different/same shuffle seeds and sequence length of the teacher during inference. The last column shows the performance of the offline (cached) implementation relative to an online implementation, where the entire teacher model is run.}
\label{table: Implementation Detail Online Offline}
\end{table}

\section{Downstream Evaluation Details}
\subsection{Natural Language Understanding}
\label{sec:appendix_nlu}
We evaluate the downstream natural language understanding performance of our trained models using the following benchmarks: HellaSwag~\citep{zellers_hellaswag_2019}, Arc-Easy~\citep{clark_think_2018}, LAMBADA~\citep{paperno_lambada_2016}, and PiQA~\citep{bisk_piqa_2020}. We conduct zero-shot evaluation of all benchmarks using LM-Eval-Harness~\citep{eval-harness}. In the main paper, we report the average scores obtained across these tasks, and full scores are provided in \cref{tab:all_downstream_results}.

\subsection{Supervised Finetuning for Instruction Following}

We used the Olmo2~\citep{2OLMo2} version of the Tulu~\citep{Tulu3Pushing} Instruction Following dataset for SFT training after Language Modeling pre-training.

\subsection{Instruction Following Evaluation}
\label{sec:appendix_IF}
Similar to \citet{MiniLLMKnowledgeDistillation}, we evaluate the ability of fine-tuned models to follow instructions on five datasets: 

\begin{itemize}
    \item \textbf{DollyEval}~\citep{conover_free_2023}: 15k human-written instruction-response pairs. Following \citet{MiniLLMKnowledgeDistillation}, we use the 500-sample test set for evaluation.
    \item \textbf{SelfInst}~\citep{wang_self-instruct_2023}: A user-oriented instruction following dataset containing 252 samples.
    \item \textbf{VicunaEval}~\citep{chiang_vicuna_2023}: 80 diverse and challenging question-answer pairs.
    \item \textbf{S-NI}: The test set of Supernatural Instruction~\citep{wang_super-naturalinstructions_2022}. We sample 1694 pairs whose ground-truth response length is longer than 11.
    \item \textbf{UnNI}: A 10k subset of Unnatural Instruction~\citep{honovich_unnatural_2023}. Similar to S-NI, we only use pairs where the ground-truth length is longer than 11.
\end{itemize}

We adopt the LLM-as-a-Judge approach, where we use Llama 3.1 405B Instruct~\cite{grattafiori2024llama3herdmodels} to score the quality of model responses. For each instruction, we generate the response five times using different seeds and temperature = 1. We prompt the judge model to rate both the ground-truth response and the model-generated response on a scale of 1-10, and use the average ratio of the total score of the ground-truth and model-generated responses as the final score. 

\section{Hyper-parameters}
\label{sec:appendix_hparams}

The hyper-paramerters for our experiments are described in \cref{hparams_300M,hparams_llama,hparams_llama_sft,hparams_arch}
\begin{table}[ht]
\begin{center}
\begin{tabular}{lc}
\hline
\toprule
\textbf{Parameters} & \textbf{Values} \\
\toprule
Optimizer & Adam \\
$\beta_1, \beta_2$ & $0.9, 0.95$ \\
Effective Batch Size & $1024$ \\
Drop-out ($p$) & $0.0$ \\
Sequence Length & $1024$ \\
Train Iters & $10,000 $ \\
Learning rate & $4*10^{-4}$ \\
Schedule & Cosine / Constant \\
LR Decay Iterations & $100\%$ \\
Warmup steps & $4\%$  \\ 
Min LR & $4*10^{-5}$ \\
Gradient clipping & $1.0$ \\
\bottomrule
\end{tabular}
\end{center}
\caption{Pre-Training Hyper-Parameters for $300$M model. The pre-training dataset was web data, primarily Fineweb-Edu.}
\label{hparams_300M}
\end{table}

\begin{table}[ht]
\begin{center}
\begin{tabular}{lc}
\hline
\toprule
\textbf{Parameters} & \textbf{Values} \\
\toprule
Optimizer & Adam \\
$\beta_1, \beta_2$ & $0.9, 0.95$ \\
Effective Batch Size & $1024$ \\
Drop-out ($p$) & $0.0$ \\
Sequence Length & $4096$ \\
Train Iters & $10,000 $ \\
Learning rate & $3*10^{-4}$ \\
Schedule & Cosine  \\
LR Decay Iterations & $100\%$ \\
Warmup steps & $4\%$  \\ 
Min LR & $3*10^{-5}$ \\
Gradient clipping & $1.0$ \\
\bottomrule
\end{tabular}
\label{hparams_llama}
\end{center}
\caption{Training Hyper-Parameters for $3$B Llama model}
\end{table}

\begin{table}[ht]
\begin{center}
\begin{tabular}{lc}
\hline
\toprule
\textbf{Parameters} & \textbf{Values} \\
\toprule
Optimizer & Adam \\
$\beta_1, \beta_2$ & $0.9, 0.95$ \\
Effective Batch Size & $256$ \\
Drop-out ($p$) & $0.0$ \\
Sequence Length & $4096$ \\
Train Iters & $1,234 $ \\
Learning rate & $2*10^{-5}$ \\
Schedule & Cosine  \\
LR Decay Iterations & $100\%$ \\
Warmup steps & $3\%$  \\ 
Min LR & $2*10^{-6}$ \\
Gradient clipping & $1.0$ \\
\bottomrule
\end{tabular}
\end{center}
\caption{SFT Hyper-Parameters for $3$B Llama model}
\label{hparams_llama_sft}
\end{table}

\begin{table}[ht]
\begin{center}
\begin{adjustbox}{max width=\linewidth}
\begin{tabular}{lcc}
\hline
\toprule
\textbf{Parameters} & \textbf{300M Model} & \textbf{3B Model}\\
\toprule
Num Layers  & $24$ & $28$ \\
Hidden Size & $1024$ & $3072$ \\
FFN Hidden Size & $2816$ & $8192$ \\
Num Attn Heads & $8$ & $24$ \\
Num Query Groups & $8$/$4$ & $8$ \\
\bottomrule
\end{tabular}
\end{adjustbox}
\end{center}
\caption{Student Model Architecture Details. The $100$B experiments for $300$M model used $4$ query groups for efficiency. The pre-training dataset was FineWeb-Edu~\cite{finewebdatasetsdecantingweb}}
\label{hparams_arch}
\end{table}


\section{Package versions}

Versions of packages used are described in \cref{package}.

\begin{table}[ht]
\begin{center}
\begin{tabular}{lc}
\hline
\toprule
\textbf{Package} & \textbf{Version} \\
\toprule
megatron & $0.7.0$ \\ 
deepspeed & $0.15.3$ \\ 
flash\_attn & $2.4.2$ \\ 
safetensors & $0.4.5$ \\ 
scikit-learn & $1.5.2$ \\ 
scipy & $1.14.0$ \\ 
sentencepiece & $0.2.0$ \\ 
torch & $2.5.0$ \\ 
transformer\_engine & $1.11.0$ \\ 
transformers & $4.46.1$ \\
\bottomrule
\end{tabular}
\end{center}
\caption{Package Versions for Pre-training}
\label{package}
\end{table}

\section{Computational Resources}
All experiments were carried out on nodes with 8 Nvidia H100 GPUs with $80$Gb memory. Most experiments utilized one node or less, while the large scale ones used $2-4$ nodes.

\section{Use of AI Assistants}
AI assistants were consulted while writing a small fraction of the code for this work, but their work was carefully checked, and the majority of the code was handwritten. AI assistants were not used in writing the text of this paper.

\section{Artifacts}
We use LLaMA-3-8B~\citep{grattafiori2024llama3herdmodels} as the teacher for some of experiments. We also used the Llama-3.1-405b as a judge for evaluation. Both of these uses are permitted under the license of these models. The datasets used here are also permitted for research use, and were only used for research. The pre-training dataset Fineweb-Edu~\citep{finewebdatasetsdecantingweb} is primarily composed of English educational-style web data, and so is the SFT data Tulu~\cite{Tulu3Pushing}.

\clearpage
\onecolumn
\section{Pseudo-code}

\definecolor{LightGray}{gray}{0.9}

The pseudocode for topk sampling and random sampling approaches is provided below.

\DeclareRobustCommand{\ttfamily}{\fontencoding{T1}\fontfamily{lmtt}\selectfont}

\begin{minted}
[
frame=lines,
framesep=3mm,
baselinestretch=1,
fontsize=\tiny,
rulecolor=\color{gray!80},
bgcolor=LightGray,
]{python}
import torch

## Create downsampled probabilities
def create_prob(values, indices, probs):
    downsampled_probs = torch.zeros_like(probs)
    downsampled_probs.scatter_(1, indices, values)
    return downsampled_probs

## Downsampling Functions
def downsample_topk(probs, k=50):  # Top-k
    topk_values, topk_indices = probs.topk(k)
    return create_prob(topk_values, topk_indices, probs)

def downsample_ours(probs, N=50):  # Sampling
    sampled_indices = torch.multinomial(probs, N, replacement=True)
    prob_value = 1.0 / N
    values = torch.full((probs.size(0), N), prob_value, device=probs.device)
    return create_prob(values, sampled_indices, probs)

## Knowledge distillation loss
def distillation_loss(student_logits, teacher_probs, downsample_fn):
    # Downsample teacher distribution
    downsampled_teacher_probs = downsample_fn(teacher_probs)
    
    # Compute KL divergence
    loss =  torch.nn.functional.kl_div(
         torch.nn.functional.log_softmax(student_logits, dim=-1),
        downsampled_teacher_probs,
    )
    return loss

## Training step
def train_step(inputs, labels, teacher_model, student_model, downsample_fn, alpha=0.5):
    # Forward pass through teacher and student
    with torch.no_grad():
        teacher_logits = teacher_model(inputs)
        teacher_probs =  torch.nn.functional.softmax(teacher_logits, dim=-1)
    
    student_logits = student_model(inputs)
    
    # Compute standard cross-entropy loss
    ce_loss =  torch.nn.functional.cross_entropy(student_logits, labels)
    
    # Compute distillation loss
    kd_loss = distillation_loss(student_logits, teacher_probs, downsample_fn)
    
    # Combine losses
    total_loss = alpha * kd_loss + (1 - alpha) * ce_loss
    
    return total_loss
\end{minted}

\onecolumn

\definecolor{LightGray}{gray}{0.9}

The pseudocode for running different sampling strategies on a toy distribution. 

\DeclareRobustCommand{\ttfamily}{\fontencoding{T1}\fontfamily{lmtt}\selectfont}

\begin{minted}
[
frame=lines,
framesep=3mm,
baselinestretch=1,
fontsize=\tiny,
rulecolor=\color{gray!80},
bgcolor=LightGray,
]{python}
# Set random seed for reproducibility
np.random.seed(12345)

# Configuration parameters
VOCAB_SIZE = 100000
TOP_K = 20
NUM_SAMPLES = 22
NUM_SAMPLING_ROUNDS = 1000
Y_MAX = 50

# Create synthetic data distribution
def create_synthetic_data(vocab_size):
    idx = np.array(range(1, vocab_size + 1))
    data_dist = 1 / idx
    data_dist /= np.sum(data_dist)  # Normalize to sum to 1
    return idx, data_dist

# Generate data
idx, data_dist = create_synthetic_data(VOCAB_SIZE)

# Top-K method
def apply_top_k(data_dist, idx, top_k):
    top_k_probs = data_dist[:top_k]
    top_k_probs_redistributed = top_k_probs / np.sum(top_k_probs)
    # top_k_probs_redistributed = top_k_probs
    
    # Create top-k distribution with a small offset for visualization
    top_k_dist = np.zeros_like(data_dist)
    top_k_dist[:top_k] = top_k_probs_redistributed
    top_k_dist = list(top_k_dist[:top_k]) + [0] + list(top_k_dist[top_k:])
    return top_k_dist

data_dist_top_k = apply_top_k(data_dist, idx, TOP_K)

# Naive fix method
def apply_naive_fix(data_dist, idx, top_k):
    naive_fix_dist = np.zeros_like(data_dist)
    naive_fix_dist[:top_k] = data_dist[:top_k]
    naive_fix_dist += data_dist * (1 - np.sum(naive_fix_dist))
    return naive_fix_dist

data_dist_remaining_gt = apply_naive_fix(data_dist, idx, TOP_K)

# Random sampling method
def apply_random_sampling(data_dist, num_samples, num_rounds):
    random_sampling_dist = np.zeros_like(data_dist)
    num_samples_effective = 0
    
    for _ in range(num_rounds):
        current_dist = np.zeros_like(data_dist)
        samples = np.random.choice(len(data_dist), size=num_samples, p=data_dist)
        for i in samples:
            current_dist[i] += 1
        num_samples_effective += np.count_nonzero(current_dist)
        current_dist /= num_samples
        random_sampling_dist += current_dist
    
    num_samples_effective /= num_rounds
    random_sampling_dist /= np.sum(random_sampling_dist)
    return random_sampling_dist, num_samples_effective

data_dist_random_sampling, num_samples_effective = apply_random_sampling(data_dist, NUM_SAMPLES, NUM_SAMPLING_ROUNDS)


def plot_probability_distributions(LINE_WIDTH=2.0, MARKER_SIZE=3):
    plt.plot(idx[:Y_MAX], data_dist[:Y_MAX], label='Ground Truth', color='purple', linewidth=LINE_WIDTH, marker='o', markersize=MARKER_SIZE)
    
    # Plot Top-K distribution
    idx_topk = list(idx[:TOP_K]) + list(idx[TOP_K:])
    data_dist_top_k_truncated = list(data_dist_top_k[:TOP_K])  + list(data_dist_top_k[TOP_K:])
    plt.plot(idx_topk[:Y_MAX+1], data_dist_top_k_truncated[:Y_MAX+1], 
                 label='Top-K (k=20)', color='royalblue', linewidth=LINE_WIDTH, marker='o', markersize=MARKER_SIZE)
    
    plt.plot(idx[:Y_MAX], data_dist_remaining_gt[:Y_MAX], 
                 label='Naive Fix', color='darkgoldenrod', linewidth=LINE_WIDTH, marker='o', markersize=MARKER_SIZE)
    
    plt.plot(idx[:Y_MAX], data_dist_random_sampling[:Y_MAX], 
                 label='Random Sampling', color='salmon', linewidth=LINE_WIDTH, marker='o', markersize=MARKER_SIZE)
    
    # Add plot details
    plt.ylim(-0.002, 0.15)
    plt.legend(fontsize=12, framealpha=0.6)
    plt.xticks(fontsize=11)
    plt.yticks(fontsize=11)
    plt.grid()
    plt.xlabel(r'Token Index $\rightarrow$', fontsize=14)
    plt.ylabel(r'Teacher Probability $\rightarrow$', fontsize=14)    
    plt.savefig("images/image.png", dpi=600, bbox_inches='tight')

plot_probability_distributions()
print(f"Effective number of samples: {num_samples_effective:.2f}")
\end{minted}
\label{toy_prob_pseduo}

\onecolumn

\definecolor{LightGray}{gray}{0.9}

The pseudocode for running different top-k strategies on a synthetic classification task. 

\DeclareRobustCommand{\ttfamily}{\fontencoding{T1}\fontfamily{lmtt}\selectfont}

\begin{minted}
[
frame=lines,
framesep=3mm,
baselinestretch=1,
fontsize=\tiny,
rulecolor=\color{gray!80},
bgcolor=LightGray,
]{python}

torch.random.manual_seed(1234)
torch.set_default_dtype(torch.float64)
device='cuda'
num_classes = 1024
sigma = 1.5
num_dim = 128
num_hidden_teacher = 128
num_hidden_student = 96
class_centers = torch.rand((num_classes, num_dim), device=device)
class_sigma = torch.unsqueeze(torch.rand((num_classes, ), device=device), dim=-1) * sigma
class_indices = torch.tensor(range(num_classes), device=device)
num_calibration_batches = 100

def get_batch(batch_size=4096):
    idx = torch.randint(low=0, high=num_classes, size=(batch_size,), device=device)
    class_centers_batch = class_centers[idx]
    class_sigma_batch = class_sigma[idx]
    batch = class_centers_batch + torch.randn((batch_size, num_dim), device=device)*class_sigma_batch
    return batch, idx

def eval(model, method):
    all_probs = []
    all_acc = []
    with torch.no_grad():
        for i in tqdm(range(num_calibration_batches)):
            model.eval()
            batch, labels = get_batch()
            probs = model(batch)
            probs = torch.nn.functional.softmax(probs, dim=-1)
            all_probs.append(torch.max(probs, dim=-1)[0])
            all_acc.append(torch.argmax(probs, dim=-1).detach() == labels)
    all_probs = torch.vstack(all_probs)
    all_acc = torch.vstack(all_acc)
    print(f'Accuracy for {method}', all_acc.float().mean().item()*100)

def train(model, method, teacher=None, lr=2e-3, num_rounds=20000, **kwargs):
    optimizer = torch.optim.AdamW(params = model.parameters(), lr=lr, weight_decay=0.00)
    for step in tqdm(range(num_rounds)):
        optimizer.zero_grad()
        batch, labels = get_batch()
        logits = model(batch)
        if teacher:
            teacher.eval()
            logits_teacher = teacher(batch)
            probs_teacher = torch.nn.functional.softmax(logits_teacher, dim=-1).detach()
            loss = loss_kd(logits, probs_teacher, method, **kwargs)
        else:
            loss = torch.nn.functional.cross_entropy(logits, labels)
        loss.backward()
        optimizer.step()
    eval(model, method)
    return model


def loss_kd(logits, probs_teacher, method, topk=7, to_sample=50):
    if "topk" in method:
        topk_probs, topk_ids = probs_teacher.topk(topk, dim=-1)
        probs_teacher *= 0
        probs_teacher.scatter_reduce_(dim=-1, index=topk_ids, src=topk_probs, reduce='sum')
    elif "random_sampling" in method:
        probs_teacher_cumsum = probs_teacher.cumsum(dim=-1)  
        rand_probs = torch.rand(size=(probs_teacher_cumsum.shape[0], to_sample), device=probs_teacher_cumsum.device)
        rand_probs = rand_probs.sort(dim=-1)[0]
        sample_token_ids = torch.searchsorted(probs_teacher_cumsum, rand_probs) # Inverse Transform Sampling
        probs_teacher *= 0
        probs_teacher.scatter_reduce_(dim=-1, index=sample_token_ids, src=torch.ones_like(probs_teacher), reduce='sum')
        probs_teacher.div_(probs_teacher.sum(dim=-1, keepdim=True))

    logits_exp = torch.exp(logits)
    logits_sum_exp = torch.sum(logits_exp, dim=-1)
    logits_log_sum_exp = torch.log(logits_sum_exp)
    loss = - probs_teacher * (logits - torch.unsqueeze(logits_log_sum_exp, dim=-1))
    loss = torch.sum(loss, dim=-1).mean()
    return loss

class ToyModel(torch.nn.Module):
    def __init__(self, num_hidden):
        super().__init__()
        self.layer1 = torch.nn.Linear(num_dim, num_hidden)
        self.layer2 = torch.nn.Linear(num_hidden, num_hidden)
        self.layer3 = torch.nn.Linear(num_hidden, num_classes)
    def forward(self, x):
        x = torch.nn.functional.gelu(self.layer1(x))
        x = torch.nn.functional.gelu(self.layer2(x))
        x = self.layer3(x)
        return x

teacher = train(ToyModel(num_hidden_teacher).to(device), 'teacher')
student = train(ToyModel(num_hidden_student).to(device), 'student')
student_kd = train(ToyModel(num_hidden_student).to(device), 'student_full_kd', teacher=teacher)
student_topk = train(ToyModel(num_hidden_student).to(device), 'student_topk', teacher=teacher, topk=7)
student_random = train(ToyModel(num_hidden_student).to(device), 'student_random_sampling', teacher=teacher, to_sample=50)
\end{minted}
\label{toy_example_pseduo}

\section{NLU Tasks Full Scores}
\begin{table}[ht]
\centering
\begin{adjustbox}{max width=\linewidth}
\begin{tabular*}{\textwidth}{l@{\extracolsep{\fill}} c c c c c | c}
\toprule
\textbf{Experiment} & \textbf{ARC Easy} & \textbf{HellaSwag} & \textbf{LAMBADA} & \textbf{LAMBADA} & \textbf{PIQA} & \textbf{Avg.} \\
\textbf{} & \textbf{} & \textbf{} & \textbf{OpenAI} & \textbf{Standard} & \textbf{} & \textbf{} \\
\midrule
\multicolumn{7}{c}{\textbf{3B Teacher $\rightarrow$ 300M Student}} \\
\midrule
\multicolumn{7}{l}{\textbf{Base}} \\
CE          & 46.59 & 41.18 & 38.85 & 30.80 & 67.41 & 44.97 \\
Ours (12)   & 50.76 & 41.84 & 40.25 & 30.70 & 67.46 & 46.20 \\
FullKD      & 51.56 & 41.98 & 40.69 & 29.52 & 67.25 & 46.20 \\
\midrule
\midrule
\multicolumn{7}{c}{\textbf{8B Teacher $\rightarrow$ 3B Student}} \\
\midrule
\multicolumn{7}{l}{\textbf{Base}} \\
CE                      & 64.90 & 56.35 & 45.64 & 38.31 & 72.58 & 55.56 \\
Top12                   & 65.07 & 57.04 & 47.76 & 39.86 & 73.50 & 56.65 \\
Top50                   & 65.66 & 57.80 & 47.88 & 40.87 & 73.50 & 57.14 \\
Ours (12)               & 66.29 & 58.93 & 47.47 & 40.99 & 73.83 & 57.50 \\
Ours (12)++             & 68.14 & 60.82 & 46.83 & 39.80 & 73.99 & 57.92 \\
FullKD                  & 66.08 & 58.76 & 48.01 & 40.71 & 73.88 & 57.49 \\
\midrule
\multicolumn{7}{l}{\textbf{SFT, Tulu}} \\
CE                      & 58.84 & 57.51 & 45.66 & 37.92 & 72.69 & 54.52 \\
Top12                   & 63.51 & 58.49 & 50.92 & 42.97 & 72.47 & 57.67 \\
Top50                   & 66.58 & 59.26 & 50.86 & 42.07 & 72.91 & 58.34 \\
Ours (12)               & 68.43 & 60.14 & 52.14 & 42.67 & 73.83 & 59.44 \\
Ours (12)++             & 66.96 & 60.91 & 50.71 & 42.23 & 74.48 & 59.06 \\
FullKD                  & 68.22 & 59.59 & 50.46 & 42.32 & 73.01 & 58.72 \\
\bottomrule
\end{tabular*}
\end{adjustbox}
\caption{Full performance results on various benchmarks for $300$M and $3$B experiments.}
\label{tab:all_downstream_results}
\end{table}

\end{document}